\documentclass[11pt]{article}

\usepackage{latexsym}

\newcommand{\inn}{\hbox{\bf in}}
\newcommand{\out}{\hbox{\bf out}}
\newcommand{\lar}{\leftarrow}
\newcommand{\lacute}{\langle}
\newcommand{\racute}{\rangle}
\newcommand{\cln}{\!\!:\!\!}
\newcommand{\pcomp}{\hbox{\it pcomp}}

\newtheorem{theorem}{Theorem}[section]
\newtheorem{lemma}[theorem]{Lemma}

\newtheorem{corollary}[theorem]{Corollary}
\newtheorem{example}[theorem]{Example}
\newtheorem{definition}[theorem]{Definition}

\newtheorem{remark}[theorem]{Remark}

\begin{document}

\pagestyle{plain}

\title{Annotated revision programs}

\author{Victor Marek \and Inna Pivkina \and Miros\l aw Truszczy\'nski}

\date{}

\maketitle

\begin{center}
Department of Computer Science,\\
University of Kentucky,\\
Lexington, KY 40506-0046\\
{\tt marek|inna|mirek@cs.engr.uky.edu}
\end{center}

\begin{abstract}
Revision programming is a formalism to describe and enforce updates of 
belief sets and databases. That formalism was extended by Fitting who 
assigned annotations to revision atoms. Annotations provide a way to 
quantify the confidence (probability) that a revision atom holds. The 
main goal of our paper is to reexamine the work of Fitting, argue that 
his semantics does not always provide results consistent with intuition, 
and to propose an alternative treatment of annotated revision programs. 
Our approach differs from that proposed by Fitting in two key aspects: 
we change the notion of a model of a program and we change the notion 
of a justified revision. We show that under this new approach 
fundamental properties of justified revisions of standard revision 
programs extend to the annotated case.\\

\end{abstract}

\section{Introduction}

Revision programming is a formalism to specify and enforce constraints
on databases, belief sets and, more generally, on arbitrary sets.
Revision programming was introduced and studied in
\cite{mt95a,mt94c}. The formalism was shown to be closely related
to logic programming with stable model semantics \cite{mt94c,pt97}.
In \cite{mpt98}, a simple correspondence of revision programming with
the general logic programming system of Lifschitz and Woo \cite{lw92}
was discovered. Roots of another recent formalism of dynamic logic
programming \cite{alppp98} can also be traced back to revision
programming.
 
(Unannotated) revision rules come in two forms of {\em in-rules} and
{\em out-rules}:
\begin{equation}
\label{in.eq.1}
\inn(a)\lar\inn(a_1),\ldots,\inn(a_m),\out(b_1),\ldots,\out(b_n)
\end{equation}
and
\begin{equation}
\label{in.eq.2}
\out(a)\lar\inn(a_1),\ldots,\inn(a_m),\out(b_1),\ldots,\out(b_n).
\end{equation}
Expressions $\inn(a)$ and $\out(a)$ are called {\em revision atoms}.
Informally, the atom $\inn(a)$ stands for ``$a$ is in the current set''
and $\out(a)$ stands for ``$a$ is not in the current set.''
The rules (\ref{in.eq.1}) and (\ref{in.eq.2}) have the following
interpretation: whenever all elements $a_k$, $1\leq k\leq m$,
belong to the current set (database, belief set) and none of the elements
$b_l$, $1\leq l\leq n$, belongs to the current set then, in the case of rule
(\ref{in.eq.1}), the item $a$ should be in the revised set, and in the
case of rule (\ref{in.eq.2}), $a$ should not be in the revised set.

Let us illustrate the use of the revision rules by an example.
\begin{example}\label{ex-1990}{\rm
Let program $P$ consist of the following two rules.
$$\inn(b)\leftarrow \out(c)\qquad\mbox{ and }\qquad 
\inn(c)\leftarrow \inn(a), \out(b).$$
When the current set (initial database) has only atom $a$ in it, there
are two intended revisions. One of them consists of $a$ and $b$. 
The other one consists of $a$ and $c$.
If, however, the initial database is empty, there is only one intended
revision consisting of atom $b$.
}\hfill$\Box$
\end{example}
 
To provide a precise semantics to {\em revision programs} (collections 
of revision rules), the concept of a {\em justified revision}
was introduced in \cite{mt95a,mt94c}. Informally, given an initial
set $B_I$ and a revision program $P$, a justified revision of $B_I$ with
respect to $P$ (or, simply, a $P$-justified revision of $B_I$) is obtained
from $B_I$ by adding some elements to $B_I$ and by removing some other
elements from $B_I$ so that each change is, in a certain sense, justified.
The intended revisions discussed in Example \ref{ex-1990} are $P$-justified
revisions. 
 
The formalism of revision programs was extended by Fitting \cite{fit95}
to the case when revision atoms occurring in rules are assigned
{\em annotations}. Such annotation can be interpreted as the degree
of confidence that a revision atom holds. For instance, an annotated
atom $(\inn(a)\cln 0.2)$ can be regarded as the statement that $a$ is
in the set with the probability $0.2$. Thus, annotated atoms and
annotated revision programs can be used to model situations when
membership status of atoms (whether they are ``in'' or ``out'') is not
precisely known and when constraints reflect this imprecise knowledge.
In his work, Fitting defined the concept of an annotated revision
program, described the concept of a justified revision of a database by
an annotated revision program, and studied properties of that notion.
 
The annotations do not have to be numeric. In fact they may come from
any set. It is natural, though, to assume that the set of annotations
has a mathematical structure of a complete distributive lattice. 
Such lattices allow us to capture within
a single algebraic formalism different intuitions associated with
annotations. For instance, annotations expressing probabilities
\cite{ns94}), possibilistic annotations \cite{vE86}, and annotations
in terms of opinions of groups of experts \cite{fit95} can all be
regarded as elements of certain complete and distributive lattices. 
The general formalism of lattice-based annotations was studied by 
Kifer and Subrahmanian \cite{ksub92} but only for logic programs 
without negations.
 
In the setting of logic programs, an annotation describes the probability
(or the degree of belief) that an atom is implied by a program or, that it
is ``in'' a database. The closed world assumption then implies the
probability that an atom is ``out''. Annotations in the context of
revision programs provide us with richer descriptions of the status of
atoms. Specifically, a possible interpretation of a {\em pair} of annotated
revision literals $(\inn(a):\alpha)$ and $(\out(a):\beta)$ is that our
confidence in $a$ being in a database is $\alpha$ and that, in the same
time, our confidence that $a$ does not belong to the database is $\beta$.
Annotating atoms with {\em pairs} of annotations allows us to model
incomplete and contradictory information about the status of an atom.
 
Thus, in annotated revision programming the status of an atom $a$ is, in
fact, given by a pair of annotations. Therefore, in this paper we will
consider, in addition to a lattice of annotations, which we will denote
by ${\cal T}$, the product of ${\cal T}$ by itself --- the lattice
${\cal T}^2$. There are two natural orderings on ${\cal T}^2$. We will use
one of them, the {\em knowledge} ordering, to compare the degree of
incompleteness (or degree of contradiction) of the pair of annotations
describing the status of an atom.
 
The main goal of our paper is to reexamine the work of Fitting, argue
that his semantics does not always provide results consistent with
intuition, and to propose an alternative treatment of annotated revision
programs. Our approach differs from that proposed by Fitting in two
key aspects: we use the concept of an {\em s-model} which is a refinement of
the notion of a model of a program, and we change the notion of a
justified revision. We show that under this new approach fundamental
properties of justified revisions of standard revision programs extend
to the case of annotated revision programs.
 
Here is a short description of the content and the contributions of our
paper. In Section \ref{prelim}, we introduce annotated revision
programs, provide some examples and discuss underlying motivations. We
define the concepts of a valuation of a set of revision atoms in
a lattice of annotations ${\cal T}$ and of a valuation of a set of
(ordinary) atoms in the corresponding product lattice ${\cal T}^2$. 
We also define the knowledge ordering on ${\cal T}^2$ and on valuations 
of atoms in ${\cal T}^2$.
 
Given an annotated revision program, we introduce the notion of the
operator associated with the program. This operator acts on valuations
in ${\cal T}^2$ and is analogous to the van Emden-Kowalski operator
for logic programs \cite{vEK76}. It is monotone with respect to
the knowledge ordering and allows us to introduce the notion of
the necessary change entailed by an annotated revision program.
 
In Section \ref{intuition}, we introduce one of the two main concepts
of this paper, namely that of an s-model of a revision program. Models of
annotated revision programs may be inconsistent. In the case of an
s-model, if it is inconsistent, its inconsistencies are explicitly or 
implicitly supported by the program and the model itself. We contrast 
the notion of an s-model with that of a model. We show that in general 
the two concepts are different. However, we also show that under
the assumption of consistency they coincide.
 
In Section \ref{s4}, we define the notion of a justified revision of an
annotated database by an annotated revision program $P$. Such revisions
are referred to as $P$-justified revisions. They are defined so as to
generalize justified revisions of \cite{mt95a,mt94c}.
 
Justified revisions considered here are different from those introduced
by Fitting in \cite{fit95}. We provide examples that show that Fitting's
concept of a justified revision fails to satisfy some natural postulates
and argue that our proposal more adequately models intuitions associated
with annotated revision programs. In the same time, we provide a
complete characterization of those lattices for which both proposals
coincide. In particular, they coincide in the standard case of revision
programs without annotations.
 
We study the properties of justified revisions in Section \ref{props}.
We show that annotated revision programs with the semantics of justified
revisions generalize revision programming as introduced and studied in
\cite{mt95a,mt94c}. Next, we show that $P$-justified revisions are s-models
of the program $P$. Thus, the concept of an s-model introduced in Section
\ref{prelim} is an appropriate refinement of the notion of a model to be
used in the studies of justified revisions. Further, we prove that
$P$-justified revisions decrease inconsistency and, consequently, that
a consistent model of a program $P$ is its own unique $P$-justified
revision.
 
Throughout the paper we adhere to the syntax of annotated revision 
programs proposed by Fitting in \cite{fit95}. This syntax stems
naturally from the syntax of ordinary revision programs introduced
in \cite{mt95a,mt94c} and allows us to compare directly our approach
with that of Fitting. However, in Section \ref{shift}, we propose and
study an alternative syntax for annotated revision programs. In this new
syntax (ordinary) atoms are annotated by elements of the product lattice 
${\cal T}^2$. Using this alternative syntax, we obtain an elegant 
generalization of the shifting theorem of \cite{mpt98}. 

In Section \ref{concl}, we provide a brief account of some miscellaneous
results on annotated revision programs. In particular, we discuss the
case of programs with disjunctions in the heads and the case when the
lattice of annotations is not distributive.

\section{Preliminaries}\label{prelim}

We will start with examples that illustrate main notions and a possible
use of annotated revision programming. Formal definitions will follow.

\begin{example}\label{proposal}{\rm 

A group of experts is about to discuss a certain proposal and then vote 
whether to accept or reject it.
Each person has an opinion on the proposal that may
be changed during the discussion as follows:
\begin{itemize}
\item[-] any person can convince an optimist
to vote for the proposal,
\item[-] any person can convince a pessimist to vote
against the proposal.
\end{itemize}
The group consists of two optimists (Ann and Bob) and
one pessimist (Pete). We want to be able to answer the following question:
given everybody's opinion on the subject before the
discussion, what are the possible outcomes of the vote?

Assume that before the vote Pete is for the proposal, Bob is against, 
and Ann is indifferent (has no arguments for and no arguments against the 
proposal). This situation can be described by assigning to atom 
``$accept$'' the annotation $\lacute \{Pete\}, \{Bob\}\racute$, where 
the first element of the pair is the set of experts who have arguments 
for the acceptance of the proposal and the second element is the set of 
experts who have arguments against the proposal. In the formalism of 
annotated revision programs, as proposed by Fitting in \cite{fit95}, 
this initial situation is described by a {\em function} that assigns to 
each atom in the language (in this example there is only one atom) 
its annotation. In our example, this function is given by:
$B_I(accept) = \langle\{Pete\},\{Bob\}\rangle$. (Let us mention here that 
in general, the sets of experts in an annotation need not to be disjoint. 
An expert may have arguments for and against the proposal at the same 
time. In such a case the expert is contradictory.)

The ways in which opinions may change are described by the following 
annotated revision rules:
\begin{eqnarray*}
(\inn(accept)\cln\{Ann\}) & \leftarrow & (\inn(accept)\cln\{Bob\}) \cr
(\inn(accept)\cln\{Ann\}) & \leftarrow & (\inn(accept)\cln\{Pete\}) \\
(\inn(accept)\cln\{Bob\}) & \leftarrow & (\inn(accept)\cln\{Ann\}) \cr
(\inn(accept)\cln\{Bob\}) & \leftarrow & (\inn(accept)\cln\{Pete\}) \\
(\out(accept)\cln\{Pete\}) & \leftarrow & (\out(accept)\cln\{Ann\}) \\
(\out(accept)\cln\{Pete\}) & \leftarrow & (\out(accept)\cln\{Bob\})
\end{eqnarray*} 
The first rule means that if Bob accepts the proposal, then Ann should 
accept the proposal, too, since she will be convinced by Bob. Similarly, 
the second rule means that if Pete has arguments for the proposal, then
he will be able to convince Ann. These two rules describe Ann being an 
optimist. The remaining rules follow as Bob is an optimist and Pete is 
a pessimist.

Possible outcomes of the vote are given by justified revisions.
In this particular case there are two justified revisions of the 
initial database $B_I$. 
They are $B_R(accept) = \langle\{Ann, Bob, Pete\},$ $\{\}\rangle$
and $B_R'(accept) = \langle\{\},\{Bob, Pete\}\rangle$.
The first one corresponds to the case when the proposal is accepted
(Ann, Bob and Pete all voted for). This outcome happens if Pete 
convinces Bob and Ann to vote for.
The second revision corresponds to the case when Bob and Pete voted 
against the proposal (Ann remained indifferent and did not vote).
This outcome happens if Bob convinces Pete to change 
his opinion.
}\hfill$\Box$
\end{example}

\begin{remark}
It is possible to rewrite annotated revision rules from Example 
\ref{proposal} as ordinary revision rules (without annotations)
if we use atoms "$accept\_Ann$'', "$accept\_Bob$'', and "$accept\_Pete$''.
However, ordinary revision programs do not deal with inconsistent or 
not completely defined databases. In particular, we will not be able to 
express the fact that initially Ann has no arguments for and no 
arguments against the proposal in Example \ref{proposal}. 

\end{remark}

In the next example annotations are real numbers from the interval $[0,1]$
representing different degrees of a particular quality.

\begin{example}\label{lights}{\rm 

Assume that there are two sources of light: $a$ and $b$.
Each of them may be either On or Off.
They are used to transmit two signals.
The first signal is a combination of $a$ being On and $b$ being Off.
The second signal is a combination of $a$ being Off and $b$ being On.

The sources $a$ and $b$ are located far from an observer. Such factors as
light pollution and dust may affect the perception of signals.
Therefore, the observed brightness of a light source differs from its
actual brightness.
Assume that brightness is measured on a scale from 0 (complete darkness)
to 1 (maximal brightness). The actual brightness of a light source
may be either $0$ (when it is Off), or $1$ (when it is On).

Initial database $B_I$ represents observed brightness of sources.
For example, if observed brightness of source $a$ is $\alpha$
($0\leq\alpha\leq 1$), then $B_I(a)=\lacute\alpha,1-\alpha\racute$.
We may think of the first and the second elements in the
pair $\lacute\alpha,1-\alpha\racute$
as degrees of brightness and darkness of the source respectfully.
The task is to infer actual brightness from observed brightness.
Thus, revision of the initial database should represent actual brightness
of sources.

Suppose we know that dust in the air cannot reduce brightness by
more than $0.2$. Then, we can safely assume that a light source is On if its
observed brightness is $0.8$ or more.
Assume also that light pollution cannot contribute more than
$0.4$. That is, if observed darkness of a source is at least $0.6$, it must be
Off. This information together with the fact that only two signals are
possible, may be represented by the following annotated revision program $P$:
\begin{eqnarray*}
(\inn(a)\cln 1) & \leftarrow & (\inn(a)\cln 0.8), (\out(b)\cln 0.6) \cr
(\out(b)\cln 1) & \leftarrow & (\inn(a)\cln 0.8), (\out(b)\cln 0.6) \cr
(\inn(b)\cln 1) & \leftarrow & (\inn(b)\cln 0.8), (\out(a)\cln 0.6) \cr
(\out(a)\cln 1) & \leftarrow & (\inn(b)\cln 0.8), (\out(a)\cln 0.6)
\end{eqnarray*}
The first two rules state that if the brightness of $a$ is at least $0.8$ and
darkness of $b$ is at least $0.6$, then brightness of $a$ is $1$
(the first rule)
and darkness of $b$ is $1$ (the second rule). This corresponds to the case
when the first signal is transmitted. Similarly, the last two rules describe
the case when the second signal is transmitted.

Let observed brightness of $a$ and $b$ be $0.3$ and $0.9$ respectively.
That is, $B_I(a)=\lacute 0.3,0.7\racute$ and $B_I(b)=\lacute 0.9, 0.1\racute$.
Then, $P$-justified revision of $B_I$ is the actual brightness. In this
case we have
$B_R(a)=\lacute 0,1\racute$ ($a$ is Off), and 
$B_R(b)=\lacute 1,0\racute$ ($b$ is On).
}\hfill$\Box$
\end{example}

Now let us move on to formal definitions. Throughout the paper we 
consider a fixed {\em universe} $U$ whose elements are referred to 
as {\em atoms}. In Example \ref{proposal} $U=\{accept\}$. In Example
\ref{lights} $U=\{a,b\}$. Expressions of 
the form $\inn(a)$ and $\out(a)$, where $a\in U$, are called {\em 
revision atoms}. In the paper we assign annotations to revision atoms. 
These annotations are members of a {\em complete infinitely distributive 
lattice 
with the De Morgan complement} (an order reversing involution). 
Throughout the paper this lattice is denoted by $\cal T$. The partial 
ordering on $\cal T$ is denoted by $\leq$ and the corresponding meet 
and join operations by $\wedge$ and $\vee$, respectively. The De Morgan 
complement of $a \in {\cal T}$ is denoted by $\overline{a}$. Let us 
recall that it satisfies the following two laws (the De Morgan laws):
$$\overline{a\vee b} = \overline{a} \wedge \overline{b}, \qquad 
\overline{a\wedge b} = \overline{a}\vee \overline{b}.$$
In Example \ref{proposal}, $\cal T$ is the set of subsets of the set 
$\{Ann, Bob, Pete\}$, with $\subseteq$ as the ordering relation, and
the set-theoretic complement as the De Morgan complement.
In Example \ref{lights}, ${\cal T}= [0,1]$ with the usual ordering; 
the De Morgan complement of $\alpha$ is $1-\alpha$.
 
An {\em annotated revision atom} is an expression of the form
$(\inn(a)\cln \alpha)$ or $(\out(a)\cln \alpha)$, where $a\in U$ and 
$\alpha\in {\cal T}$. An {\em annotated revision rule} is an expression 
of the form
\[
p\leftarrow q_1,\dots,q_n,
\]
where $p$, $q_1,\dots,q_n$ are annotated revision atoms. An {\em annotated
revision program} is a set of annotated revision rules.
 
A {\em ${\cal T}$-valuation} is a mapping from the set of revision atoms
to ${\cal T}$. A ${\cal T}$-valuation $v$ describes our information
about the membership of the elements from $U$ in some (possibly unknown)
set $B\subseteq U$. For instance, $v(\inn(a)) =\alpha$ can be
interpreted as saying that $a\in B$ with certainty $\alpha$.
A ${\cal T}$-valuation $v$ {\em satisfies}
an annotated revision atom $(\inn(a)\cln \alpha)$ if $v(\inn(a))\geq\alpha$.
Similarly, $v$ {\em satisfies} $(\out(a)\cln \alpha)$ if $v(\out(a))\geq\alpha$.
The ${\cal T}$-valuation $v$ {\em satisfies} a list or a set of annotated
revision atoms if it satisfies each member of the list or the set.
A $\cal T$-valuation {\em satisfies} an annotated revision rule if it
satisfies the head of the rule whenever it satisfies the body of the rule.
Finally, a $\cal T$-valuation {\em satisfies} an annotated revision
program (is a {\em model} of the program) if it satisfies all rules in
the program.
 
Given an annotated revision program $P$ we can assign to it an operator on the
set of all $\cal T$-valuations. Let $t_{P}(v)$ be the set of the heads
of all rules in $P$ whose bodies are satisfied by  a $\cal T$-valuation $v$. 
We define an operator $T_P$ as follows:
\[
T_P(v)(l) = \bigvee\{\alpha|(l\cln \alpha)\in t_P(v)\}
\]
Here $\bigvee X$ is the join of the subset $X$ of the lattice
(note that $\bot$ is the join of an empty set of lattice elements).
The operator $T_P$ is a counterpart of the well-known van Emden-Kowalski
operator from logic programming and it will play an important role in
our paper.
 
It is clear that under $\cal T$-valuations, the information about an
element $a\in U$ is given by a pair of elements from $\cal T$ that are
assigned to revision atoms $\inn(a)$ and $\out(a)$. Thus, in the paper
we will also consider an algebraic structure ${\cal T}^2$ with
the domain ${\cal T}\times {\cal T}$ and with an ordering $\leq_k$
defined by:
\[
\lacute \alpha_1,\beta_1 \racute \leq_k \lacute \alpha_2,\beta_2
\racute\ \ \ \mbox{if}\ \ \ \alpha_1\leq\alpha_2 \ \mbox{and} \
\beta_1\leq\beta_2.
\]
If a pair $\lacute \alpha_1,\beta_1 \racute$ is viewed as a measure
of our information about membership of $a$ in some unknown set $B$
then $\alpha_1\leq\alpha_2$ and $\beta_1 \leq\beta_2$ imply that
the pair $\lacute \alpha_2,\beta_2 \racute$ represents higher degree of
knowledge about $a$. Thus, the ordering $\leq_k$ is often referred to as
the {\em knowledge} or {\em information} ordering. Since the lattice
$\cal T$ is complete and distributive, ${\cal T}^2$ is a complete 
distributive lattice with respect to the ordering $\leq_k$\footnote{There 
is another ordering that can be associated
with ${\cal T}^2$. We can define $\lacute \alpha_1,\beta_1 \racute \leq_t
\lacute \alpha_2,\beta_2 \racute$ if $\alpha_1\leq\alpha_2$ and
$\beta_1\geq\beta_2$. This ordering is often called the {\em truth
ordering}. Since $\cal T$ is a complete distributive lattice, 
${\cal T}^2$ with both
orderings $\leq_k$ and $\leq_t$ forms a 
complete distributive {\em bilattice} (see \cite{gin88,fi99}
for a definition). In this paper we will not use the ordering $\leq_t$
nor the fact that ${\cal T}^2$ is a bilattice.}.
 
The operations of meet, join, top, and bottom under $\leq_k$ are denoted
$\otimes$, $\oplus$, $\top$, and $\perp$, respectively. In addition, we
make use of the {\em conflation} operation. Conflation is defined as
$-\lacute \alpha,\beta \racute = \lacute \overline{\beta},\overline{\alpha}
\racute$. An element $A\in {{\cal T}^2}$ is {\em consistent} if
$A\leq_k -A$. In other words, an element $\langle\alpha,\beta\rangle
\in {\cal T}^2$ is consistent if $\alpha$ is smaller than or equal to
the complement of $\beta$ (the evidence ``for'' is less than or equal
than the complement of the evidence ``against'') and $\beta$ is smaller
than or equal to the complement of $\alpha$ (the evidence ``against'' is
less than or equal than the complement of the evidence ``for'').   

The conflation operation satisfies the De Morgan laws:
$$-(\lacute \alpha,\beta \racute \oplus \lacute\gamma,\delta\racute) = 
-\lacute \alpha,\beta \racute \otimes -\lacute\gamma,\delta\racute,$$
$$-(\lacute \alpha,\beta \racute \otimes \lacute\gamma,\delta\racute) = 
-\lacute \alpha,\beta \racute \oplus -\lacute\gamma,\delta\racute,$$
where $\alpha,\beta,\gamma,\delta\in {\cal T}$.
 

A ${{\cal T}^2}$-valuation is a mapping from {\em atoms} to
elements of ${{\cal T}^2}$. If $B(a) = \lacute\alpha,\beta
\racute$ under some ${{\cal T}^2}$-valuation $B$, we say that under $B$ the
element
$a$ is in a set with certainty $\alpha$ and it is not in the set
with certainty $\beta$. We say that a ${{\cal T}^2}$-valuation is {\em
consistent} if it assigns a consistent element of ${\cal T}^2$ to every
atom in $U$.
 
In this paper, ${\cal T}^2$-valuations will be used to represent current
information about sets (databases) as well as the change that needs to be
enforced. Let $B$ be a ${\cal T}^2$-valuation representing our knowledge
about a certain set and let $C$ be a ${\cal T}^2$-valuation representing
change that needs to be applied to $B$. We define the revision
of $B$ by $C$, say $B'$, by
\[
B'=(B\otimes -C)\oplus C.
\]
The intuition is as follows. After the revision, the new valuation must
contain at least as much knowledge about atoms being in and out as $C$.
On the other hand, this amount of knowledge must not exceed implicit
bounds present in $C$ and expressed by $-C$, unless $C$ directly
implies so. In other words, if $C(a) = \lacute
\alpha,\beta\racute$, then evidence for $\inn(a)$ must not exceed
$\bar{\beta}$ unless $\alpha\geq \bar{\beta}$, 
and the evidence for $\out(a)$ must not exceed
$\bar{\alpha}$ unless $\beta\geq \bar{\alpha}$. Since we prefer
explicit evidence of $C$ to implicit evidence expressed by $-C$, we
perform the change by first using $-C$ and then applying $C$. However,
let us note here that the order matters only if $C$ is inconsistent; if
$C$ is consistent, $(B\otimes -C)\oplus C = (B\oplus C)\otimes -C$.
This specification of how the change modeled by a ${\cal T}^2$-valuation
is enforced plays a key role in our definition of justified revisions in
Section \ref{s4}. 

\begin{example}[continuation of Example \ref{proposal}]{\rm 
In Example \ref{proposal}, $B_I$ has two revisions.
The first one, $B_R$, is the 
revision of $B_I$ by $C$, where
$C(accept)=\langle\{Ann, Bob\},\{\}\rangle$. 
We have $-C(accept)=\langle\{Ann, Bob, Pete\},\{Pete\}\rangle$.
Thus, $(B_I\otimes -C)(accept) = \langle\{Pete\},\emptyset\rangle$, and
$((B_I\otimes -C)\oplus C)(accept) = 
\langle\{Ann, Bob, Pete\},\emptyset\rangle = B_R(accept)$.

The second revision, $B_R'$, is the revision of $B_I$ by $C'$, where
$C'(accept)=\langle\{\},\{Pete\}\rangle$.
}\hfill$\Box$
\end{example}
 
There is a one-to-one correspondence
$\theta$ between ${\cal T}$-valuations (of revision atoms) and
${{\cal T}^2}$-valuations (of atoms). For a ${\cal T}$-valuation
$v$, the ${{\cal T}^2}$-valuation $\theta(v)$ is defined by:
$\theta(v)(a)= \lacute v(\inn(a)),v(\out(a))\racute$.
The inverse mapping of $\theta$ is denoted by $\theta^{-1}$.
Clearly, by using the mapping $\theta$, the notions of satisfaction defined
earlier for ${\cal T}$-valuations can be extended to ${\cal T}^2$-valuations.
Similarly, the operator $T_P$ gives rise to a related
operator $T^b_P$. The operator $T_P^b$ is defined on the set of
all ${\cal T}^2$-valuations by
$T^b_P = \theta\circ T_P\circ \theta^{-1}$. The key property of the
operator $T_P^b$ is its $\leq_k$-monotonicity.
 
\begin{theorem}\label{kmono}
Let $P$ be an annotated revision program and let $B$ and $B'$ be
two ${\cal T}^2$-valuations such that $B\leq_k B'$.
Then, $T_P^b(B)\leq_k T_P^b(B')$.
\end{theorem}
 
By Tarski-Knaster Theorem \cite{ta58} it follows that the operator $T_P^b$ has
a least fixpoint in ${\cal T}^2$ (see also \cite{ksub92}). This fixpoint is an
analogue of the concept
of a least Herbrand model of a Horn program. It represents the set of
annotated revision atoms that are implied by the program and, hence,
must be satisfied by any revision under $P$ of {\em any} initial valuation.
Given an annotated revision program $P$ we will refer to the least
fixpoint of the operator $T_P^b$ as the {\em necessary change} of $P$ and
will denote it by $NC(P)$. The present concept of the necessary change
generalizes the corresponding notion introduced in \cite{mt95a,mt94c}
for the original unannotated revision programs.
 
To illustrate concepts and results of the paper, we will consider
two special lattices.
The first of them is the lattice with the domain $[0,1]$
(interval of reals), with the standard ordering $\leq$, and the standard
complement operation $\bar{\alpha}= 1-\alpha$. 
We will denote this lattice by ${\cal T}_{[0,1]}$.
Intuitively, the annotated revision atom $(\inn(a)\cln x)$, where $x\in [0,1]$,
stands for the statement that $a$ is ``in'' with likelihood (certainty)
$x$.
 
The second lattice is the Boolean algebra of all subsets of a given set
$X$. It will be denoted by ${\cal T}_X$. We will think of elements from
$X$ as experts. The annotated revision atom $(\out(a)\cln Y)$, where
$Y\subseteq X$, will be understood as saying that $a$ is believed to be
``out'' by those experts that are in $Y$ (the atom
$(\inn(a)\cln Y)$ has a similar meaning).

\section{Models and s-models}\label{intuition}
 
The semantics of annotated revision programs will be based on
the notion of a model, as defined in the previous section, and on its
refinements. The first two results describe some simple properties of
models of annotated revision programs. The first of them characterizes
models in terms of the operator $T^b_P$.
 
\begin{theorem}\label{model}
Let $P$ be an annotated revision program. A ${\cal T}^2$-valuation 
$B$ is a model of $P$ (satisfies $P$) if and only if $B \geq_k 
T_{P}^b(B)$.
\end{theorem}

Models of annotated revision programs are closed under meets. This
property is analogous to a similar property holding for models of Horn
programs. Indeed, since $B_1 \otimes B_2 \leq_k B_i$, $i=1,2$, and $T_P^b$ 
is $\leq_k$-monotone, by Theorem \ref{model} we obtain
\[
T_P^b(B_1 \otimes B_2) \leq_k T_P^b(B_i) \leq_k B_i, \ \ i=1,2.
\]
Consequently, 
\[
T_P^b(B_1 \otimes B_2) \leq_k B_1 \otimes B_2.
\]
Thus, again by Theorem \ref{model} we obtain the following result.
 
\begin{corollary}\label{meetofmodels}
The meet of two models of an annotated revision program $P$ is 
also a model of $P$.
\end{corollary}
 
Given an annotated revision program $P$, its necessary change $NC(P)$
satisfies $NC(P)=T_P^b(NC(P))$. Hence, $NC(P)$ is a model of $P$.

As we will now argue, not all models are appropriate for describing
the meaning of an annotated revision program. The problem is that
${\cal T}^2$-valuations may contain inconsistent information about elements
from $U$. When studying the meaning of an annotated revision program
we will be interested in those models only whose inconsistencies are
limited to those explicitly or implicitly supported by 
the program and by the model itself.
 
Consider the program $P=\{(\inn(a)\cln \{q\}) \lar\}$ (where the
annotation $\{q\}$ comes from the  lattice
${\cal T}_{\{p,q\}}$). This program
asserts that $a$ is ``in'', according to expert $q$. By closed world
assumption, it also implies an upper bound for the evidence for
$\out(a)$. In this case the only expert that might possibly believe 
in $\out(a)$ is $p$ (this is to say that expert $q$ does
not believe in $\out(a)$). 
Observe that a ${\cal T}^2$-valuation $B$, such that
$B(a) = \lacute \{q\},\{q\}\racute$ is a model of $P$
but it does not satisfy the implicit bound on evidence for $\out(a)$.

Let $P$ be an annotated program and let $B$ be a ${\cal T}^2$-valuation
that is a model of $P$. 
By the explicit evidence we mean evidence provided by heads of
program rules applicable with respect to $B$, that is with bodies 
satisfied by $B$. 
It is $T^b_P(B)$.
The implicit information is given by
a version of the closed world assumption: if the maximum evidence
for a revision atom $l$ provided by the program is $\alpha$ then,
the evidence for the dual revision atom $l^D$ ($\out(a)$,
if $l=\inn(a)$, or $\inn(a)$, otherwise) must not exceed
$\bar{\alpha}$ (unless explicitly forced by the program).
Thus, the implicit evidence is given by $-T^b_P(B)$.
Hence, a model $B$ of a program $P$ contains no more
evidence than what is directly implied by $P$ given $B$
and what is indirectly implied by $P$ given $B$
if $B\leq_k T^b_P(B) \oplus (-T^b_P(B))$ (since the direct
evidence is given by $T_P^b(B)$ and the implicit evidence is given by
$-T_P^b(B)$). This observation leads us to a refinement of the notion
of a model of an annotated revision program.
 
\begin{definition}
Let ${P}$ be an annotated revision program and let $B$ be a
${{\cal T}^2}$-valuation. We say that $B$ is an {\em s-model} of $P$ if
\[
T_{P}^b(B) \leq_k B \leq_k T_{P}^b(B)\oplus (-T_{P}^b(B)).
\]
\end{definition}

The ``s" in the term ``s-model" stands for ``supported'' and emphasizes 
that inconsistencies in s-models are limited to those explicitly or 
implicitly supported by the program and the model itself.
 
Clearly, by Theorem \ref{model}, an s-model of $P$ is a model of $P$.
In addition, it is easy to see that the necessary change of an annotated
program $P$ is an s-model of $P$ (it follows directly from the fact that
$NC(P)=T_P^b(NC(P)))$.
 
The distinction between models and s-models appears only in the context
of inconsistent information. This observation is formally stated below.
 
\begin{theorem}\label{cons_mod}
Let $P$ be an annotated revision program. A consistent ${\cal T}^2$-valuation
$B$ is an s-model of $P$ if and only if $B$ is a model of $P$.
\end{theorem}
Proof. $(\Rightarrow)$
Let $B$ be an s-model of $P$. Then,
$T_{P}^b(B) \leq_k B \leq_k
T_{P}^b(B)\oplus (-T_{P}^b(B))$. In particular,
$T_{P}^b(B) \leq_k B$ and, by Theorem \ref{model}, $B$ is a model of
$P$. \\
$(\Leftarrow)$
Let $B$ satisfy $P$. From Theorem \ref{model} we have
$T_{P}^b(B)\leq_k B$. Hence, $-B \leq_k -T_{P}^b(B)$.
Since $B$ is consistent, $B\leq_k  -B$. Therefore,
\begin{equation}\label{eqqq-1}
T_{P}^b(B)\leq_k B\leq_k  -B\leq_k -T_{P}^b(B).
\end{equation}
It follows that
$T_{P}^b(B)\leq_k -T_{P}^b(B)$ and
$T_{P}^b(B)\oplus (-T_{P}^b(B)) = -T_{P}^b(B)$.
By (\ref{eqqq-1}), we get
\[
T_{P}^b(B)\leq_k B\leq_k T_{P}^b(B)\oplus (-T_{P}^b(B))
\]
and the assertion follows. \hfill$\Box$

Some of the properties of ordinary models hold for s-models, too.
For instance, the following theorem shows that an s-model of two 
annotated revision programs is 
an s-model of their union.

\begin{theorem}\label{cmodunion}
Let $P_1$, $P_2$ be annotated revision programs.
Let $B$ be an s-model of $P_1$ and an s-model of $P_2$.
Then, $B$ is an s-model of $P_1\cup P_2$.
\end{theorem}
Proof. 
Clearly, $B$ is a model of $P_1\cup P_2$. That is,
\begin{equation}\label{eqv-1}
T_{P_1\cup P_2}^b(B) \leq_k B.
\end{equation}
It is easy to see that 
\begin{equation}\label{eqv-178}
T_{P_1\cup P_2}^b(B) = T_{P_1}^b(B) \oplus T_{P_2}^b(B).
\end{equation}
Hence, by the De Morgan law,
\begin{equation}\label{eqv-179}
-T_{P_1\cup P_2}^b(B) = -T_{P_1}^b(B) \otimes -T_{P_2}^b(B).
\end{equation}
It follows from the definition of an s-model that
$$B \leq_k T_{P_1}^b(B)\oplus -T_{P_1}^b(B) 
\mbox{, and }$$
$$B \leq_k T_{P_2}^b(B)\oplus -T_{P_2}^b(B).$$
Thus,
$$B \leq_k (T_{P_1}^b(B)\oplus -T_{P_1}^b(B))\otimes
(T_{P_2}^b(B)\oplus -T_{P_2}^b(B)).$$
By the distributivity of lattice operations in ${\cal T}^2$, we obtain
$$B \leq_k (T_{P_1}^b(B)\otimes (T_{P_2}^b(B)\oplus -T_{P_2}^b(B)))\oplus
(-T_{P_1}^b(B)\otimes (T_{P_2}^b(B)\oplus -T_{P_2}^b(B))).$$
The first summand is smaller or equal to 
$T_{P_1}^b(B)$. Thus, by applying distributivity to the second
summand, we get the following inequality:
$$B \leq_k T_{P_1}^b(B) \oplus 
(-T_{P_1}^b(B)\otimes T_{P_2}^b(B)) \oplus 
(-T_{P_1}^b(B)\otimes -T_{P_2}^b(B)).$$
Using $-T_{P_1}^b(B)\otimes T_{P_2}^b(B)\leq_k T_{P_2}^b(B)$ and then
(\ref{eqv-178}) and (\ref{eqv-179}), we get
$$B \leq_k T_{P_1}^b(B) \oplus T_{P_2}^b(B) \oplus -T_{P_1\cup
P_2}^b(B) =
T_{P_1\cup P_2}^b(B) \oplus -T_{P_1\cup P_2}^b(B).$$
In other words,
\begin{equation}\label{eqv-2}
B \leq_k T_{P_1\cup P_2}^b(B) \oplus -T_{P_1\cup P_2}^b(B).
\end{equation}
From (\ref{eqv-1}) and (\ref{eqv-2}) it follows that $B$ is an s-model 
of $P_1\cup P_2$. \hfill$\Box$

Not all of the properties of models hold for s-models.
For instance, the counterpart of Corollary \ref{meetofmodels} does not hold.
The following example shows that the meet of two s-models 
is not necessarily an s-model.
\begin{example}{\rm 
Consider the lattice ${\cal T}_{\{p,q\}}$.
Let $P$ be an annotated program consisting of the following rules:
\begin{eqnarray*}
(\inn(a)\cln \{p\}) & \leftarrow & (\inn(b)\cln \{p\}) \\ 
(\out(a)\cln \{p\}) & \leftarrow & \\
(\inn(a)\cln \{p\}) & \leftarrow & (\out(b)\cln \{p\})
\end{eqnarray*}
Let $B_1$ and $B_2$ be defined as follows.
$$B_1(a) =\lacute\{p\},\{p\}\racute,\qquad 
B_1(b) =\lacute\{p\},\emptyset\racute;$$
$$B_2(a) =\lacute\{p\},\{p\}\racute,\qquad 
B_2(b) =\lacute\emptyset,\{p\}\racute.$$
Let us show that $B_1$ is an s-model of $P$. Indeed,
$$T_{P}^b(B_1)(a) = \lacute\{p\},\{p\}\racute,\qquad
T_{P}^b(B_1)(b) = \lacute\emptyset,\emptyset\racute.$$
Hence, 
$$-T_{P}^b(B_1)(a) = \lacute\{q\},\{q\}\racute,\qquad
-T_{P}^b(B_1)(b) = \lacute\{p,q\},\{p,q\}\racute.$$
Therefore,
$$T_{P}^b(B_1)(a)\le_k B_1(a) \le_k (T_{P}^b(B_1)\oplus -T_{P}^b(B_1))(a),
\mbox{ and}$$
$$T_{P}^b(B_1)(b)\le_k B_1(b) \le_k (T_{P}^b(B_1)\oplus -T_{P}^b(B_1))(b).$$
In other words, $B_1$ is an s-model of $P$. 
Similarly, $B_2$ is an s-model of $P$.
However, $B_1\otimes B_2$ is {\em not} an s-model of $P$.
Indeed,
$$(B_1\otimes B_2)(a) = \lacute\{p\},\{p\}\racute,\qquad
(B_1\otimes B_2)(b) = \lacute\emptyset,\emptyset\racute.$$
Then,
$$T_{P}^b(B_1\otimes B_2)(a) = \lacute\emptyset,\{p\}\racute,\qquad
T_{P}^b(B_1\otimes B_2)(b) = \lacute\emptyset,\emptyset\racute, \mbox{ and}$$
$$-T_{P}^b(B_1\otimes B_2)(a) = \lacute\{q\},\{p,q\}\racute,\qquad
-T_{P}^b(B_1\otimes B_2)(b) = \lacute\{p,q\},\{p,q\}\racute.$$
Hence,
$$(B_1\otimes B_2)(a) \not\leq_k 
(T_{P}^b(B_1\otimes B_2)\oplus -T_{P}^b(B_1\otimes B_2))(a) = 
\lacute\{q\},\{p,q\}\racute.$$
Therefore, $B_1\otimes B_2$ is not an s-model of $P$.
}\hfill$\Box$
\end{example}

In this example both $B_1$ and $B_2$, as well as their meet
$B_1\otimes B_2$ are inconsistent. 
For $B_1$ and $B_2$ there are rules in $P$ that explicitly 
imply their inconsistencies.
However, for $B_1\otimes B_2$ the bodies of these rules are no longer 
satisfied. Consequently, the inconsistency in $B_1\otimes B_2$ is not 
implied by $P$. That is, $B_1\otimes B_2$ is not an s-model of $P$.

Let us now investigate 
what happens when we add to an annotated revision program
$P$ a rule $r= (l\cln \alpha) \leftarrow (l\cln\alpha)$
(here $l$ is a revision atom, $\alpha$ is an annotation).
Unlike ordinary revision programs where every database is a model of
a rule of the form $l\leftarrow l$, not every ${{\cal T}^2}$-valuation
is an s-model of $r$. Therefore, adding such a rule may affect the set of
s-models of the program. On the one hand, rule $r$ by imposing
additional implicit bound on $l^D$ may give rise to a situation when
an s-model of $P$ is not an an s-model of $P\cup \{r\}$
(Case 1 of Example \ref{add_rule}).
On the other hand, rule $r$ may provide additional explicit
evidence for $l$ that results in a situation when an s-model of $P\cup \{r\}$
is not an s-model of $P$ (Case 2 of Example \ref{add_rule}).

\begin{example}\label{add_rule}{\rm
Let $U=\{a\}$ and the lattice of annotations be ${\cal T}_{\{p,q\}}$.
Let $B(a)=\lacute\{p\},\{p\}\racute$.
Let $r= \ (\inn(a)\cln\{p\})\leftarrow (\inn(a)\cln\{p\})$.
\begin{enumerate}
\item
Let $P=\{\}$.
Then, $T^b_P(B)(a) = \lacute\emptyset,\emptyset\racute$, and
$-T^b_P(B)(a) = \lacute\{p,q\},\{p,q\}\racute$.
Hence, $T^b_P(B)(a)\leq B(a)\leq T_{P}^b(B)(a)\vee (-T_{P}^b(B))(a)$.
Thus, $B$ is an s-model
of $P$. However, $B$ is {\em not} an s-model of $P\cup \{r\}$.
Indeed, $T^b_{P\cup \{r\}}(B)(a) = \lacute\{p\},\emptyset\racute$, and
$-T^b_{P\cup \{r\}}(B)(a) = \lacute\{p,q\},\{q\}\racute$.
Hence, $B(a)\not\leq T^b_{P\cup \{r\}}(B)(a)\vee (-T^b_{P\cup \{r\}}(B))(a) =
\lacute\{p,q\},\{q\}\racute$.
Therefore, $B$ is {\em not} an s-model of $P\cup \{r\}$.
\item
Let $P=\{\ (\out(a)\cln\{p\})\leftarrow \ \}$.
Then it is easy to see that $B$ is {\em not} an s-model
of $P$. However, $B$ is an s-model of $P\cup \{r\}$. \hfill$\Box$
\end{enumerate}
}
\end{example}

\begin{remark}
Let us note that adding rule $r= (l\cln \alpha) \leftarrow (l\cln\alpha)$ 
to $P$ has no effect on consistent models of
$P$. Indeed, let $B$ be a consistent model of $P$. Clearly, $B$ is
a model of $\{r\}$.
Hence, by Theorem \ref{cons_mod}, $B$ is an s-model of $P$, and 
an s-model of $\{r\}$.
Therefore, Theorem \ref{cmodunion} implies that 
$B$ is an s-model of $P\cup \{r\}$.
\end{remark}

\section{Justified revisions}\label{s4}
 
In this section, we will extend to the case of annotated revision
programs the notion of a justified revision introduced for revision
programs in \cite{mt95a}. The reader is referred to \cite{mt95a,mt94c}
for the discussion of motivation and intuitions behind the concept of
a justified revision and of the role of the {\em inertia principle}
(a version of the closed world assumption).
 
There are several properties that one would expect to hold when
the notion of justified revision is extended to the case of programs with
annotations. Clearly, the extended concept should specialize to the
original definition if annotations are dropped. Next, main
properties of justified revisions studied in \cite{mt94c,mpt98} should
have their counterparts in the case of justified revisions of annotated
programs. In particular, justified revisions of an annotated revision
program should be models of the program.

There is one other requirement
that naturally arises in the context of programs with annotations.
Consider two annotated revision rules $r$ and $r'$ that are exactly
the same except that the body of $r$ contains two annotated revision
atoms $(l\cln \beta_1)$ and $(l\cln \beta_2)$, while the body of 
$r'$ instead of $(l\cln \beta_1)$ and $(l\cln \beta_2)$ contains 
annotated revision atom $(l\cln \beta_1\vee\beta_2)$.
$$r = \qquad \dots\leftarrow \dots,(l\cln \beta_1),\dots, (l\cln
\beta_2),\dots$$
$$r' = \qquad \dots\leftarrow \dots,(l\cln \beta_1\vee\beta_2),\dots$$
We will refer to this operation as the {\em join transformation}.

It is clear, that a ${\cal T}^2$-valuation $B$ satisfies $(l\cln 
\beta_1)$ and $(l\cln \beta_2)$ if and only if $B$ satisfies 
$(l\cln \beta_1\vee\beta_2)$.
Consequently, replacing rule $r$ by rule $r'$ (or vice versa) in an
annotated revision program should have no effect on justified revisions.
In fact, any reasonable semantics for annotated revision programs should
be invariant under such operation, and we will refer to this property of
a semantics of annotated revision programs as {\em invariance under
join}.
 
Now we introduce the notion of a justified revision of an
annotated revision program and contrast it with an earlier proposal by
Fitting \cite{fit95}. In the following section we show that our concept
of a justified revision satisfies all the requirements listed above.
 
Let a ${\cal T}^2$-valuation $B_I$ represent 
our current knowledge about some subset of the universe $U$.
Let an annotated revision program $P$ describe an update that $B_I$
should be subject to.
The goal is
to identify a class of ${\cal T}^2$-valuations that could be viewed as
representing updated information about the subset, obtained by revising
$B_I$ by $P$. As argued in \cite{mt95a,mt94c}, each appropriately ``revised''
valuation $B_R$ must be {\em grounded} in $P$ and in $B_I$, 
that is, any difference
between $B_I$ and the revised ${\cal T}^2$-valuation $B_R$ must be
justified by means of the program and the information available in $B_I$.
 
To determine whether $B_R$ is grounded in $B_I$ and $P$, we use
the {\em reduct} of $P$ with respect to these two valuations. 
The construction
of the reduct consists of two steps and mirrors the original definition of
the reduct of an unannotated revision program \cite{mt94c}. In the first
step, we eliminate from $P$ all rules whose bodies are not satisfied by
$B_R$ (their use does not have
an {\em a posteriori} justification with respect to $B_R$). In the
second step, we take into account the initial valuation $B_I$.
 
How can we use the information about the initial ${\cal T}^2$-valuation
$B_I$ at this stage? Assume that $B_I$ provides evidence $\alpha$ for
a revision atom $l$. Assume also that an annotated revision atom
$(l\cln \beta)$ appears in the body of a rule $r$. In order to satisfy this
premise of the rule, it is enough to derive, from the program resulting
from step 1, an annotated revision atom $(l\cln \gamma)$, where
$\alpha\vee \gamma \geq \beta$. The least such element exists (due to
the fact that ${\cal T}$ is complete and infinitely distributive). 
Let us denote this value 
by $\pcomp(\alpha,\beta)$\footnote{The operation $\pcomp(\cdot,\cdot)$
is known in the lattice theory as the {\em relative pseudocomplement},
see \cite{rs70}.}.
 
Thus, in order to incorporate information about a revision atom $l$
contained in the initial ${\cal T}^2$-valuation $B_I$, which is given by
$\alpha = (\theta^{-1}(B_I))(l)$, we proceed as follows. In the bodies
of rules of the program obtained after step 1, we replace each annotated
revision atom of the form $(l\cln \beta)$ by the annotated revision atom
$(l\cln \pcomp(\alpha,\beta))$.
 
Now we are ready to formally introduce the notion of {\em reduct} of
an annotated revision program $P$ with respect to the pair of
${\cal T}^2$-valuations, initial one, $B_I$, and a candidate for a
revised one, $B_R$.
 
\begin{definition}\label{defreduct}
The {\em reduct} ${P}_{B_R}|B_I$ is obtained from ${P}$ by
\begin{enumerate}
\item removing every rule whose body contains an annotated
atom that is not satisfied in $B_R$,
\item replacing each annotated atom $(l\cln \beta)$
from the body of each remaining rule by the annotated atom
$(l\cln \gamma)$, where $\gamma = \pcomp ((\theta^{-1}(B_I))(l),\beta)$.
\end{enumerate}
\end{definition}
 
We now define the concept of a {\em justified revision}.
Given an annotated revision program $P$, we first compute the
reduct ${P}_{B_R}|B_I$ of the program $P$
with respect to $B_I$ and $B_R$. Next, we compute the necessary change
for the reduced program. Finally we apply this change to
the ${\cal T}^2$-valuation $B_I$. A ${\cal T}^2$-valuation $B_R$ is a
justified revision of $B_I$ if the result
of these three steps is $B_R$.  Thus we have the following definition.
 
\begin{definition}\label{defjr}
$B_R$ is a ${P}$-{\em justified revision} of $B_I$ if
$B_R = (B_I\otimes -C)\oplus C$, where $C=NC(P_{B_R}|B_I)$ is the
necessary change for ${P}_{B_R}|B_I$.
\end{definition}
 
We will now contrast  this approach with the one proposed by
Fitting in \cite{fit95}. In order to do so, we recall the definitions
introduced in \cite{fit95}. The key difference is in the way Fitting
defines the reduct of a program. The first step is the same in both
approaches. However, the second steps, in which the initial valuation is
used to simplify the bodies of the rules not eliminated in the first
step of the construction, differ.

\begin{definition}[Fitting]\label{F-rev}
Let ${P}$ be an annotated revision program and let $B_I$ and $B_R$ be
${{\cal T}^2}$-valuations. The {\em $F$-reduct} of ${P}$
with respect to $(B_I,B_R)$ (denoted ${P}^F_{B_R}|B_I$) is defined
as follows:
\begin{enumerate}
\item Remove from ${P}$ every rule whose body contains an annotated
revision atom that is not satisfied in $B_R$.
\item From the body of each remaining rule delete any annotated revision atom
that is satisfied in $B_I$.
\end{enumerate}
\end{definition}
 
The notion of justified revision as defined by Fitting differs from our
notion only in that it uses the necessary change of the $F$-reduct
(instead of the necessary change of the reduct defined above in
Definition \ref{defreduct}). We call the justified revision based on 
the notion of $F$-reduct, the {\em $F$-justified revision}.
 
In the remainder of this section we show that the notion of the $F$-justified
revision does not in general satisfy some basic requirements that we
would like justified revisions to have. In particular, $F$-justified
revisions under an annotated revision program $P$ are not always models
of $P$.
 
\begin{example}\label{notmodel}{\rm 
Consider the lattice ${\cal T}_{\{p,q\}}$. Let $P$ be a program
consisting of the following rules:
\[
(\inn(a)\cln \{p\}) \leftarrow (\inn(b)\cln \{p,q\}) \ \ \ \mbox{and}\ \ \
(\inn(b)\cln \{q\}) \leftarrow
\]
and let $B_I$ be a valuation such that $B_I(a) =
\lacute\emptyset,\emptyset\racute$ and $B_I(b) = \lacute\{p\},\emptyset
\racute$.
Let $B_R$ be a valuation given by $B_R(a) =
\lacute\emptyset,\emptyset\racute$ and $B_R(b) = \lacute\{p,q\},\emptyset
\racute$. Clearly, ${P}^F_{B_R}|B_I = {P}$, and
$B_R$ is an ${F}$-justified revision of $B_I$ (under $P$).
However, $B_R$ does not satisfy ${P}$.
}\hfill$\Box$
\end{example}
 
The semantics of $F$-justified revisions also fails to satisfy the
invariance under join property.
 
\begin{example}{\rm 
Let ${P}$ be the same revision program as before,
and let ${P'}$ consist of the rules
\[
(\inn(a)\cln \{p\}) \leftarrow (\inn(b)\cln \{p\}), (\inn(b)\cln \{q\}) \ \ \ 
\mbox{and}\
\ \
(\inn(b)\cln \{q\}) \leftarrow
\]
Let the initial valuation $B_I$ be given by
$B_I(a) = \lacute\emptyset,\emptyset\racute$ and
$B_I(b) = \lacute\{p\},\emptyset\racute$.
The only $F$-justified revision of $B_I$ (under $P$)
is a ${\cal T}^2$-valuation $B_R$, where $B_R(a) = \lacute\emptyset,
\emptyset\racute$ and $B_R(b) = \lacute\{p,q\},\emptyset\racute$.
The only $F$-justified revision of $B_I$ (under $P'$)
is a ${\cal T}^2$-valuation $B_R'$, where
$B_R'(a) = \lacute\{p\},\emptyset\racute$ and
$B_R'(b) = \lacute\{p,q\},\emptyset\racute$.
Thus, replacing in the body of a rule 
$(\inn(b)\cln \{p,q\})$ by
$(\inn(b)\cln \{p\})$ and $(\inn(b)\cln \{q\})$ 
affects $F$-justified revisions.
}\hfill$\Box$
\end{example}
 
However, in some cases the two definitions of justified revision
coincide. The following theorem provides a complete characterization of
those cases (let us recall that a lattice ${\cal T}$ is {\em linear} if 
for any two elements $\alpha,\beta\in{\cal T}$ either
$\alpha\leq \beta$ or $\beta\leq \alpha$).
 
\begin{theorem}\label{linear}
$F$-justified revisions and justified
revisions coincide if and only if the lattice ${\cal T}$ is linear.
\end{theorem}
Proof. $(\Rightarrow)$
Assume that $F$-justified revisions and justified revisions coincide for 
a lattice ${\cal T}$.
Let $\alpha, \beta\in {\cal T}$. 
We will show that either $\alpha\leq \beta$ or $\beta\leq\alpha$.
Indeed,
let ${P}$ be annotated revision program consisting of the following rules.
\[
(\inn(a)\cln \alpha) \leftarrow (\inn(b)\cln \alpha\vee\beta) \ \ \ \mbox{and}\ \ \
(\inn(b)\cln \beta) \leftarrow
\]
Let $B_I$ be given by
$B_I(a) = \lacute\bot,\bot\racute$ and
$B_I(b) = \lacute\alpha,\bot\racute$.
Let $B_R$ be given by
$B_R(a) = \lacute\alpha,\bot\racute$ and
$B_R(b) = \lacute\alpha\vee\beta,\bot\racute$.
It is easy to see that $B_R$ is a justified revision of $B_I$ 
(with respect to $P$). 
By our assumption, $B_R$ is also an $F$-justified revision of $B_I$. 
There are only two possible cases. \\
Case 1. $\alpha\vee\beta\leq\alpha$. Then, $\beta\leq\alpha$. \\
Case 2. $\alpha\vee\beta\not\leq\alpha$.
Then,
${P}^F_{B_R}|B_I = P$. Let $C=NC({P}^F_{B_R}|B_I)$. 
By the definition of the necessary change,
$$C(a)=NC({P}^F_{B_R}|B_I)(a)=NC(P)(a)= 
\cases{\lacute\bot,\bot\racute, & when $\alpha\vee\beta\not\leq\beta$\cr
\lacute\alpha,\bot\racute, & when $\alpha\vee\beta\leq\beta$}
$$
By the definition of an $F$-justified revision,
$B_R = (B_I\otimes -C)\oplus C$.
From the facts that $B_R(a)=\lacute\alpha,\bot\racute$ and
$B_I(a) = \lacute\bot,\bot\racute$ it follows that
$C(a) = \lacute\alpha,\bot\racute$. Therefore, it is the case
that $\alpha\vee\beta\leq\beta$. That is, $\alpha\leq\beta$.\\
$(\Leftarrow)$
Assume that lattice ${\cal T}$ is linear.
Then, for any $\alpha,\beta\in {\cal T}$
$$\pcomp(\alpha,\beta) = 
\cases{\bot, & when $\alpha\geq\beta$\cr
\beta & otherwise (when $\alpha<\beta$)}
$$
Let $P$ be an annotated revision program. Let $B_I$ and $B_R$ be any 
${{\cal T}^2}$-valuations.
Let us see what is the difference between ${P}_{B_R}|B_I$ and 
${P}^F_{B_R}|B_I$.
The first steps in the definitions of reduct and $F$-reduct are the same.
During the second step of the definition 
of an $F$-reduct each annotated 
atom $(l\cln\beta)$ such that $\beta \leq B_I(l)$ is deleted from bodies 
of rules. In the second step of the definition  
of the reduct such annotated 
atom is replaced by $(l\cln\bot)$.
If $\beta > B_I(l)$, then in the reduct ${P}_{B_R}|B_I$ annotated atom
$(l\cln\beta)$ is replaced by $(l\cln\pcomp(B_I(l),\beta)) = (l\cln\beta)$,
that is, it remains as it is. In the $F$-reduct, $(l\cln\beta)$ also remains 
in the bodies for $\beta > B_I(l)$.
Thus, the only difference between ${P}_{B_R}|B_I$ and ${P}^F_{B_R}|B_I$
is that bodies of the rules from ${P}_{B_R}|B_I$ may contain atoms of 
the form $(l:\bot)$, where $l\in U$, that are not present in the bodies of 
the corresponding rules in ${P}^F_{B_R}|B_I$.
However, annotated atoms of the form $(l:\bot)$ are always satisfied.
Therefore, the necessary changes of ${P}_{B_R}|B_I$ and ${P}^F_{B_R}|B_I$,
as well as justified and $F$-justified revisions of $B_I$ coincide.
\hfill$\Box$
 
Theorem \ref{linear} explains why the difference between the justified
revisions and $F$-justified revisions is not seen when we limit our
attention to revision programs as considered in \cite{mt94c}.
Namely, the lattice ${\cal TWO} = \{{\bf f},{\bf t}\}$ of boolean values is
linear. Similarly, the lattice of reals from the
segment $[0,1]$ is linear, and there the differences cannot be seen
either.
 
\section{Properties of justified revisions}\label{props}
 
In this section we study basic properties of justified revisions. We
show that key properties of justified revisions in the case of revision
programs without annotations have their counterparts in the case of
justified revisions of annotated revision programs.
 
First, we observe that revision programs as defined in \cite{mt95a}
can be encoded as annotated revision programs (with annotations taken
from the lattice ${\cal TWO} = \{{\bf f},{\bf t}\}$). Namely, a revision
rule
\[
p\lar q_1,\ldots q_m
\]
(where $p$ and all $q_i$'s are revision atoms)
can be encoded as
\[
(p\cln {\bf t})\lar (q_1\cln {\bf t}),\ldots,(q_m\cln {\bf t})
\]
We will denote by $P^a$ the result of applying this transformation to 
a revision program $P$ (rule by rule). Second, let us represent a set of
atoms $B$ by a ${\cal TWO}^2$-valuation $B^v$ as follows:
$B^v(a)=\langle {\bf t},{\bf f}\rangle$, if $a\in B$, and
$B^v(a)=\langle {\bf f},{\bf t}\rangle$, otherwise.

Fitting \cite{fit95} argued that under such encodings the semantics 
of $F$-justified revisions generalizes the semantics of justified 
revisions introduced in \cite{mt95a}. Since for lattices whose ordering 
is linear the approach by Fitting and the approach presented in this paper
coincide, and since the ordering of ${\cal TWO}$ is linear, the semantics 
of justified revisions discussed here extends the semantics of justified
revisions from \cite{mt95a}. Specifically, we have the following result.

\begin{theorem}\label{a-2}
Let $P$ be an ordinary revision program and let $B_I$ and $B_R$ be two
sets of atoms. Then, $B_R$ is a $P$-justified revision of $B_I$ if and 
only if the necessary change of $P^a_{B_R^v}|B_I^v$ is consistent and
$B_R^v$ is a $P^a$-justified revision of $B_I^v$.
\end{theorem}

Before we study how properties of justified revisions generalize to 
the case with annotations, we prove the following auxiliary results.

\begin{lemma}\label{reductII}
Let $P$ be an annotated revision program.
Let $B$ be a ${\cal T}^2$-valuation.
Then, $NC(P_B|B) = T_{P}^b(B)$.
\end{lemma}
Proof.
The assertion follows from definitions of a necessary change and
operator $T_P^b$.
\hfill$\Box$

\begin{lemma}\label{satisf}
Let $P$ be an annotated revision program.
Let $B_I$, $B_R$, and $C$ be ${\cal T}^2$-valuations, such that
$B_R\leq B_I\oplus C$.
Then, $C$ satisfies the bodies of all rules in $P_{B_R}|B_I$.
\end{lemma}
Proof.
Let $r'\in P_{B_R}|B_I$. Let $(l\cln \gamma)$ be an annotated revision
atom
from the body of $r'$.
Let $(\theta^{-1}(B_I))(l) = \alpha$.
By the definition of the reduct, $r'$ was obtained from some rule $r\in
P$,
such that the body of $r$ is satisfied by $B_R$, and
$\gamma= \pcomp(\alpha,\beta)$,
where $(l\cln \beta)$
is in the body of $r$.
Since the body of $r$ is satisfied by $B_R$, we have
$\beta\leq (\theta^{-1}(B_R))(l)$.
From $B_R\leq_k B_I\oplus C$ it follows that
$$(\theta^{-1}(B_R))(l)\leq (\theta^{-1}(B_I\oplus C))(l) =$$
$$=(\theta^{-1}(B_I))(l)\vee (\theta^{-1}(C))(l) =
\alpha\vee (\theta^{-1}(C))(l).$$
Combining this inequality with our previous observation that
$\beta\leq (\theta^{-1}(B_R))(l)$,
we get $\beta\leq \alpha \vee (\theta^{-1}(C))(l)$. By the definition of
$\pcomp(\alpha,\beta)$, we get $\gamma\leq (\theta^{-1}(C))(l)$.
That is, $C$ satisfies $(l:\gamma)$.
Since $(l:\gamma )$ was arbitrary, $C$ satisfies all annotated
revision atoms in the body of $r'$.
As $r'$ was an arbitrary rule from $P_{B_R}|B_I$, we conclude that
$C$ satisfies the bodies of all rules in $P_{B_R}|B_I$.
\hfill$\Box$

\begin{lemma}\label{nes_cha}
Let $B_R$ be a ${P}$-justified revision of $B_I$.
Then, $NC(P_{B_R}|B_I) = T_{P}^b(B_R)$.
\end{lemma}
Proof.
By the definition of a justified revision
$B_R = (B_I\otimes -C)\oplus C$, where $C = NC(P_{B_R}|B_I)$.
Hence, $B_R\leq B_I\oplus C$. By Lemma \ref{satisf},
$C$ satisfies the bodies of all rules in $P_{B_R}|B_I$. Since $C$
is a model of $P_{B_R}|B_I$, $C$ satisfies all heads of clauses in
$P_{B_R}|B_I$.

Let $D$ be a valuation satisfying all heads of rules in $P_{B_R}|B_I$.
Then $D$ is a model of $P_{B_R}|B_I$. Since $C$ is the least
model of the reduct $P_{B_R}|B_I$, we find that $C \leq_k D$.
Consequently, $C$ is the least valuation that satisfies all heads
of the rules in $P_{B_R}|B_I$. The rules in $P_{B_R}$ are all those
rules
from $P$ whose bodies are satisfied by $B_R$. Thus, by the definition of
the
operator $T^b_P$, $C = T_{P}^b(B_R)$. \hfill$\Box$

We will now look at properties of the semantics of justified revisions. 
We will present a series of results generalizing properties of revision 
programs to the case with annotations. We will show that the concept of 
an s-model is a useful notion in the investigations of justified 
revisions of annotated programs. 

Our first result relates justified revisions to models
and s-models. Let us recall that in the case of revision programs without
annotations, justified revisions under a revision program $P$ are models
of $P$. In the case of annotated revision programs we have an analogous
result.
 
\begin{theorem}\label{isamodel}
Let ${P}$ be an annotated revision program and let $B_I$ and $B_R$ be
${\cal T}^2$-valuations. If $B_R$ is a $P$-justified revision of $B_I$
then $B_R$ is an s-model of $P$ (and, hence, a model of $P$).
\end{theorem}
Proof. 
By the definition of a $P$-justified revision,
$B_R = (B_I\otimes -C)\oplus C$, where $C$ is the necessary change
for $P_{B_R}|B_I$.
From Lemma \ref{nes_cha} it follows that $C=T_P^b(B_R)$.
Therefore,
$$B_R = (B_I\otimes -T_P^b(B_R))\oplus T_P^b(B_R)\leq_k
-T_P^b(B_R)\oplus T_P^b(B_R).$$
Also, $$B_R = (B_I\otimes -T_P^b(B_R))\oplus T_P^b(B_R)\geq
T_P^b(B_R).$$
Hence, $B_R$ is an s-model of $P$. \hfill$\Box$
 
In the previous section we showed an example demonstrating that
$F$-justified revisions do not satisfy the property of invariance
under joins. In contrast, justified revisions in the sense of our
paper do have this property.

\begin{theorem}
Let $P_2$ be the result of simplification of an annotated revision
program $P_1$ by means of the join transformation. Then for every initial
database $B_I$, $P_1$-justified revisions of $B_I$ coincide with
$P_2$-justified revisions of $B_I$. 
\end{theorem}

The proof follows directly from the definition of $P$-justified
revisions and from the following distributivity property of 
pseudocomplement: $\pcomp(\alpha,\beta_1)\vee\pcomp(\alpha,\beta_2)
= \pcomp(\alpha,\beta_1\vee\beta_2)$.

In the case of revision programs without annotations, a model of a
program $P$ is its unique $P$-justified revision. In the case
of programs with annotations, the situation is slightly more
complicated. The next several results provide a complete description
of justified revisions of models of annotated revision programs. 
First, we characterize those models that are their own justified 
revisions. This result provides additional support for the importance 
of the notion of an s-model in the study of annotated revision programs.

\begin{theorem}\label{self}
Let a ${\cal T}^2$-valuation $B_I$ be a model of an annotated revision
program ${P}$. Then, $B_I$ is a ${P}$-justified revision of itself
if and only if $B_I$ is an s-model of ${P}$.
\label{rev_of_itself}
\end{theorem}
Proof.
Let us denote $C = NC({P}_{B_I}|B_I)$. By the definition, $B_I$
is a ${P}$-justified revision of itself if and only if $B_I =
(B_I\otimes -C)\oplus C$. Since $B_I$ satisfies ${P}$, Theorem
\ref{model}
implies that $B_I \geq_k C$. Thus, $B_I\oplus C= B_I$.  Distributivity
of the product lattice ${\cal T}^2$ implies that $(B_I\otimes -C)\oplus
C
= (B_I\oplus C)\otimes (-C\oplus C) = B_I\otimes (-C\oplus C)$.
Clearly, $B_I = B_I\otimes (-C\oplus C)$ if and only if
$B_I \leq_k (-C\oplus C)$.

By Lemma \ref{reductII}, $C = NC({P}_{B_I}|B_I) = T_{P}^b(B_I)$.
Thus, $B_I$ is a ${P}$-justified revision of itself
if and only if $B_I \leq_k T_{P}^b(B_I)\oplus (-T_{P}^b(B_I))$.
But this latter condition is precisely the one that distinguishes 
s-models among models. Thus, under the assumptions of the theorem, 
$B_I$ is a $P$-justified revision of itself if and only if it is an
s-model of $P$.  \hfill$\Box$

As we observed above, in the case of programs without annotations,
models of a revision program are their own {\em unique} justified 
revisions. This property does not hold, in general, in the case of 
annotated revision programs. In other words, s-models, if they are 
inconsistent, may have other revisions besides themselves (by Theorem 
\ref{self} they always are their own revisions).
 
The following example shows that an {\em inconsistent} 
s-model may have no revisions 
other than itself, may have only one consistent justified revision, or 
may have incomparable (with respect to the knowledge ordering) 
consistent revisions. 

\begin{example}\label{multiple_revisions}{\rm 
Let the lattice of annotations be ${\cal T}_{\{p,q\}}$.
Consider an inconsistent ${\cal T}^2$-valuation $B_I$ such that $B_I(a) =
\lacute\{q\},\{q\}\racute$. 
\begin{enumerate}
\item Consider annotated revision program $P_1$ 
consisting of the clauses:
\[
(\out(a)\cln \{q\}) \leftarrow \ \ \ \mbox{and}\ \ \
(\inn(a)\cln \{q\}) \leftarrow 
\]
It is easy to see that $B_I$ is an s-model of $P_1$ and the only 
justified revision of itself.
\item Let an annotated revision program $P_2$ 
consist of the clauses:
\[
(\out(a)\cln \{q\}) \leftarrow \ \ \ \mbox{and}\ \ \
(\inn(a)\cln \{q\}) \leftarrow (\inn(a)\cln \{q\})
\]
Clearly, $B_I$
is an s-model of $P_2$. Hence, $B_I$ is its own justified revision
(under $P_2$).
 
However, $B_I$ is not the only ${P_2}$-justified revision of $B_I$.
Consider the ${\cal T}^2$-valuation $B_R$ such that $B_R(a) =
\lacute\emptyset,\{q\}\racute$. We have ${P_2}_{B_R}|B_I =
\{(\out(a)\cln \{q\}) \leftarrow\}$. Let us denote the corresponding
necessary change, $NC({P_2}_{B_R}|B_I)$, by $C$. Then, $C(a) =
\lacute\emptyset,\{q\}\racute$. Hence, $-C=\lacute\{p\},\{p,q\}\racute$
and $((B_I\otimes -C)\oplus C)(a) = \lacute\emptyset,\{q\}\racute =
B_R(a)$. Consequently, $B_R$ is a ${P_2}$-justified revision of $B_I$.
It is the only consistent ${P_2}$-justified revision of $B_I$.

\item Let an annotated revision program $P_3$ be the following:
\[
(\inn(a)\cln \{q\}) \leftarrow (\inn(a)\cln \{q\})\ \ \ \mbox{and}\ \ \
(\out(a)\cln \{q\}) \leftarrow (\out(a)\cln \{q\})
\]
Then, $B_I$ is s-model of $P_3$ and its own $P_3$-justified revision.
In addition, it is straightforward to check that 
$B_I$ has two consistent revisions
$B_R$ and $B_R'$, where
$B_R(a) = \lacute\emptyset,\{q\}\racute$ and
$B_R'(a) = \lacute\{q\},\emptyset\racute$. The revisions $B_R$ and $B_R'$
are incomparable 
with respect to the knowledge ordering. \hfill$\Box$
\end{enumerate}
}
\end{example}
 
The same behavior can be observed in the case of programs annotated with
elements from other lattices. The following example is analogous 
to the second case in the Example \ref{multiple_revisions}, but the lattice 
is ${\cal T}_{[0,1]}$.
 
\begin{example}{\rm 
Let ${P}$ be an annotated revision program (annotations
belong to the lattice ${\cal T}_{[0,1]}$) consisting of the rules:
\[
(\out(a)\cln 1) \leftarrow\ \ \ \mbox{and}\ \ \
(\inn(a)\cln 0.4) \leftarrow (\inn(a)\cln 0.4)
\]
Let $B_I$ be a valuation such that $B_I(a) = \lacute 0.4, 1\racute$.
Then, $B_I$ is an s-model of $P$ and, hence, it is its own $P$-justified
revision.
Consider a valuation $B_R$ such that $B_R(a) = \lacute 0, 1\racute$.
We have ${P}_{B_R}|B_I = \{(\out(a)\cln 1) \leftarrow\}$. Let us denote
the necessary change $NC({P}_{B_R}|B_I)$ by $C$. Then $C(a) =
\lacute 0, 1\racute$ and $-C=\lacute 0, 1\racute$.
Thus, $((B_I\otimes -C)\oplus C)(a) = \lacute 0, 1\racute = B_R(a)$.
That is, $B_R$ is a ${P}$-justified revision of $B_I$.
}\hfill$\Box$
\end{example}
 
Note that in both examples the additional justified revision $B_R$ of
$B_I$ is smaller than $B_I$ with respect to the ordering $\leq_k$. It
is not coincidental as demonstrated by our next result.
 
\begin{theorem}
Let $B_I$ be a model of an annotated revision program ${P}$.
Let $B_R$ be a ${P}$-justified revision of $B_I$. Then,
$B_R\leq_k B_I$.
\label{rev_leq_cmod}
\end{theorem}
Proof.
By the definition of a $P$-justified revision,
$B_R = (B_I\otimes -C)\oplus C$, where $C$ is the necessary change of
$P_{B_R}|B_I$.
By the definition of the reduct $P_{B_R}|B_I$ and the fact that $B_I$ 
is a model of $P$, it follows that $B_I$ is a model of $P_{B_R}|B_I$.
The necessary change $C$ is the least fixpoint of $T_{P_{B_R}|B_I}^b$, 
therefore, $C\leq B_I$.
Hence,
$$\qquad\qquad\qquad\qquad\qquad 
B_R = (B_I\otimes -C)\oplus C\leq_k B_I\oplus C\leq_k B_I\oplus B_I = B_I.
\qquad\qquad\qquad\qquad \hfill \Box$$
 
Finally, we observe that if a {\em consistent} ${\cal T}^2$-valuation is 
a model (or an s-model; these notions coincide in the class of consistent
valuations) of a program then it is its {\em unique} justified
revision.
 
\begin{theorem}\label{OnlyModel}
Let $B_I$ be a consistent model of an annotated revision
program ${P}$. Then, $B_I$ is the only ${P}$-justified revision of itself.
\end{theorem}
Proof.
Theorem \ref{cons_mod} implies that $B_I$ is an s-model of ${P}$.
Then, from Theorem \ref{rev_of_itself} we get that $B_I$ is a
${P}$-justified revision of itself.
We need to show that there are no other ${P}$-justified revisions
of $B_I$.
 
Let $B_R$ be a ${P}$-justified revision of $B_I$.
Then,  $B_R\leq_k B_I$ (Theorem \ref{rev_leq_cmod}).
Therefore, $T_{P}^b(B_R)\leq_k T_{P}^b(B_I)$.
Hence, $-T_{P}^b(B_I)\leq_k -T_{P}^b(B_R)$.
Theorem \ref{model} implies that $B_I\geq_k T_{P}^b(B_I)$.
Thus, $-B_I\leq_k -T_{P}^b(B_I)$.
Since $B_I$ is consistent, $B_I\leq_k -B_I$.
Combining the above inequalities, we get
$$B_I\leq_k -B_I\leq_k -T_{P}^b(B_I)\leq_k -T_{P}^b(B_R).$$
That is, $B_I\leq_k -T_{P}^b(B_R)$.
Hence, $B_I\otimes -T_{P}^b(B_R) = B_I$.
 
From definition of justified revision and Lemma \ref{nes_cha},
$$B_R= (B_I\otimes -T_{P}^b(B_R))\oplus T_{P}^b(B_R) =
B_I\oplus T_{P}^b(B_R)\geq_k B_I.$$
Therefore, $B_R = B_I$. \hfill $\Box$
 
To summarize, when we consider inconsistent valuations (they appear
naturally, especially when we measure beliefs of groups of independent
experts), we encounter an interesting phenomenon. An {\em inconsistent}
valuation $B_I$, even when it is an s-model of a program, may have different
justified revisions. However, all these additional revisions must be
$\leq_k$-less inconsistent than $B_I$. In the case of consistent models this
phenomenon does not occur. If a valuation $B$ is consistent and satisfies
$P$ then it is its unique $P$-justified revision.
 
In \cite{mt94c} we proved that, in the case of ordinary revision
programs, ``additional evidence does not destroy justified revisions''.
More precisely, we proved that if $B_R$ is a $P$-justified revision of
$B_I$ and $B_R$ is a model of $P'$ then $B_R$ is a $P\cup P'$-justified
revision of $B_I$. We will now prove a generalization of this property
to the case of annotated revision programs. However, as before, we
need to replace the notion of a model with that of an s-model.

\begin{theorem}
Let $P$, $P'$ be annotated revision programs.
Let $B_R$ be a $P$-justified revision of $B_I$.
Let $B_R$ be an s-model of $P'$.
Then, $B_R$ is a $P\cup P'$-justified revision of $B_I$.
\end{theorem}
Proof.
Let $C=NC(P_{B_R}|B_I)$. Let $C'=NC((P\cup P')_{B_R}|B_I)$.
Clearly, $C\leq C'$.
By the definition of a justified revision
$B_R = (B_I\otimes -C)\oplus C$.
Hence,
$$B_R\leq B_I\oplus C \leq B_I\oplus C'.$$
By Lemma \ref{satisf} it follows that
$C'$ satisfies the bodies of all rules in $(P\cup P')_{B_R}|B_I$.
Since $C'$ is the necessary change of
$(P\cup P')_{B_R}|B_I$ we conclude that $C'$ satisfies the heads of
all rules in $(P\cup P')_{B_R}|B_I$. Reasoning as in the proof of Lemma
\ref{nes_cha} we find that $C' = T_{P\cup P'}^b(B_R)$.

By Theorem \ref{isamodel}, $B_R$ is an s-model of $P$. Therefore,
by Theorem \ref{cmodunion}, $B_R$ is a s-model of $P\cup P'$.
Theorem \ref{self} implies that $B_R$ is a $P\cup P'$-justified
revision of itself. In other words,
$$B_R=(B_R\otimes -NC((P\cup P')_{B_R}|B_R))\oplus NC((P\cup
P')_{B_R}|B_R).$$
From Lemma \ref{reductII} it follows that
$NC((P\cup P')_{B_R}|B_R)= T_{P\cup P'}^b(B_R)$.
Hence, $$B_R=(B_R\otimes -C')\oplus C'.$$
Next, let us recall that $B_R = (B_I\otimes -C)\oplus C$. Hence,
$$B_R=(((B_I\otimes -C)\oplus C)\otimes -C')\oplus C'.$$
Now, using the facts that $C\leq C'$ and $-C'\leq -C$, we get
the following equalities:
$$B_R=(((B_I\otimes -C)\oplus C)\otimes -C')\oplus C' =$$
$$=((B_I\otimes -C)\otimes -C')\oplus (C\otimes -C')\oplus C' =$$
$$=(B_I\otimes (-C\otimes -C'))\oplus C' =
(B_I\otimes -C')\oplus C'$$
Thus, $B_R=(B_I\otimes -C')\oplus C'$.
By the definition of justified revisions, $B_R$ is a $P\cup
P'$-justified revision of $B_I$.
\hfill$\Box$

In case of revision programs without annotations,
justified revisions satisfy the minimality principle (see \cite{mt94c}).
Namely, $P$-justified revisions of a database differ from the database
by as little as possible. Recall, that in the case of revision programs
without annotations, databases are sets of atoms, and the difference
between databases $R$ and $I$ is their symmetric difference
$R\div I = (R\setminus I)\cup(I\setminus R)$. The minimality principle states
that if $R$ is a $P$-justified revision of $I$, then, $R\div I$ is minimal
in the family $\{B\div I : B$ is a model of $P\}$
(Theorem 3.6 in  \cite{mt94c}).

Before generalizing the minimality principle to the case of
annotated revision programs
we need to specify what we mean by the difference between
${{\cal T}^2}$-valuations.

\begin{definition}
Let $R$, $B$ be ${{\cal T}^2}$-valuations.
We say that $B$ {\em can be transformed into} $R$ {\em via} a
${{\cal T}^2}$-valuation $C$ if $R=(B\otimes -C)\oplus C$.
We say that $B$ {\em can be transformed} into $R$ if there exists
${{\cal T}^2}$-valuation $C$ such that
$B$ can be transformed into $R$ via $C$.
\end{definition}

Given two ${{\cal T}^2}$-valuations, it is not necessarily the case that
one of them can be transformed into the other. Indeed, let $V_{\top}$ be
a ${{\cal T}^2}$-valuation that assigns to each atom annotation $\top$.
Let $V_{\bot}$ be a ${{\cal T}^2}$-valuation that assigns to each
atom annotation $\bot$.
Then, if a lattice consists of more than one element, then we have
$\top\not=\bot$, and $V_{\top}$
cannot be transformed into $V_{\bot}$.

\begin{definition}
Let $R$, $B$ be ${{\cal T}^2}$-valuations.
Let $S=\{C \ |\ B$ can be transformed into $R$ via $C\}$.
The {\em difference} $\mbox{diff}(R,B)$ is
$$\mbox{diff}(R,B) =
\cases{ \prod S, & when $S\not=\emptyset$,\cr
V_{\top} & otherwise (when $S=\emptyset$).}
$$
\end{definition}

The following lemma describes a useful property of a difference between
${{\cal T}^2}$-valuations.
Namely, the difference between ${{\cal T}^2}$-valuations
$R$ and $B$ is the least (in $\leq_k$ ordering) ${{\cal T}^2}$-valuation
among all $C$ such that $R=(B\otimes -C)\oplus C$.

\begin{lemma}
Let $R$, $B$ be ${{\cal T}^2}$-valuations.
Let $S=\{C \ |\ B$ can be transformed into $R$ via $C\}$.
If $S \not = \emptyset$,
then $\mbox{diff}(R,B) \in S$.
\end{lemma}
Proof.
Let $S=\{C \ |\ B$ can be transformed into $R$ via $C\}\not = \emptyset$.
Then, $\mbox{diff}(R,B) = \prod S$.
First, let us show that $-\prod S = \sum \{-C : C\in S\}$.
On the one hand, $\prod S\leq C$ for all $C\in S$.
Thus, $-\prod S\geq -C$ for all $C\in S$.
Hence,
\begin{equation}\label{l-eq-1}
-\prod S\geq \sum \{-C : C\in S\}.
\end{equation}
On the other hand, $\sum \{-C : C\in S\}\geq -C$ for all $C\in S$.
Thus, $-\sum \{-C : C\in S\}\leq C$ for all $C\in S$.
Hence, $-\sum \{-C : C\in S\}\leq \prod S$. That is,
\begin{equation}\label{l-eq-2}
\sum \{-C : C\in S\}\geq -\prod S.
\end{equation}
From (\ref{l-eq-1}) and (\ref{l-eq-2}) it follows that
$-\prod S = \sum \{-C : C\in S\}$.

Since ${\cal T}$ is complete and infinitely distributive,
we get the following.
$$(B\otimes -\prod S)\oplus \prod S =
(B\otimes \sum \{-C : C\in S\})\oplus \prod S =$$
$$=\sum \{(B\otimes -C) : C\in S\} \oplus \prod S =$$
$$=\prod \{\sum \{(B\otimes -C) : C\in S\} \oplus C' : C'\in S\} \geq$$
$$\geq \prod \{(B\otimes -C')\oplus C' : C'\in S\} = \prod \{R\} = R.$$
That is,
\begin{equation}\label{l-eq-3}
(B\otimes -\prod S)\oplus \prod S \geq R.
\end{equation}

By definition of $S$, for each $C\in S$,
$R=(B\otimes -C)\oplus C$.
Therefore, for each $C\in S$, $C\leq R$ and $B\otimes -C\leq R$.
Thus, $\prod S\leq R$ and
$$B\otimes -\prod S = B\otimes \sum \{-C : C\in S\} =
\sum \{(B\otimes -C) : C\in S\}\leq R.$$
Hence, $(B\otimes -\prod S)\oplus \prod S\leq R$.
This together with (\ref{l-eq-3}) imply that
$$(B\otimes -\prod S)\oplus \prod S = R.$$
That is, $\prod S \in S$.
\hfill$\Box$

Now we will show that the minimality principle can be generalized
to the case of annotated revision programs.
We will have, however, to assume that ${\cal T}$ is a Boolean algebra and
restrict ourselves to consistent ${{\cal T}^2}$-valuations.

Let ${\cal T}$ be a Boolean algebra with De Morgan complement being
the complement.
Let us define the {\em negation} operation on ${\cal T}^2$ as
$\neg \lacute \alpha,\beta \racute =
\lacute \overline{\alpha}, \overline{\beta}\racute$
($\alpha,\beta\in{\cal T}$).
Then, the lattice ${\cal T}^2$ with operations $\oplus$, $\otimes$, $\neg$,
and elements $\bot$, $\top$ is a Boolean algebra, too.
Operations on ${\cal T}^2$ lift pointwise to the
space of ${\cal T}^2$-valuations. It is easy to see that
the space of ${\cal T}^2$-valuations with operations
$\oplus$, $\otimes$, $\neg$, and elements
$V_{\bot}$, $V_{\top}$ is again a Boolean algebra.

\begin{lemma}\label{inert}
Let ${\cal T}$ be a Boolean algebra.
Let $R$, $B$, $I$ be ${{\cal T}^2}$-valuations.
Let $R$ and $I$ be consistent.
Let $\mbox{diff}(R,B)\leq_k \mbox{diff}(R,I)$.
Then, $R\otimes B\geq_k R\otimes I$.
\end{lemma}
Proof.
Let $C=\mbox{diff}(R,I)$, $C'=\mbox{diff}(R,B)$.
Since $I$ is consistent,
$I\leq_k -I$. Thus,
\begin{equation}\label{equ-1}
I\otimes -(\neg I)\leq_k
-I\otimes -(\neg I)=
-(I\oplus\neg I)=
-V_{\top}=
V_{\bot}
\end{equation}
Since $R$ is consistent, $C$ is consistent, too. That is,
$C\leq_k -C$. Hence,
\begin{equation}\label{equ-2}
I\otimes -C=(I\otimes -C)\oplus(I\otimes C)
\end{equation}
Consider valuation $C\otimes \neg I$.
Using (\ref{equ-1}) and (\ref{equ-2}) we get:
$$(I\otimes -(C\otimes \neg I))\oplus (C\otimes \neg I) =
(I\otimes (-C\oplus -(\neg I)))\oplus (C\otimes \neg I) =$$
$$=(I\otimes -C)\oplus (I\otimes -(\neg I))\oplus (C\otimes \neg I)=
(I\otimes -C)\oplus(I\otimes C)\oplus V_{\bot}\oplus (C\otimes \neg I)=$$
$$=(I\otimes -C)\oplus (I\otimes C)\oplus (C\otimes \neg I)=
(I\otimes -C)\oplus (C\otimes (I\oplus\neg I))=$$
$$=(I\otimes -C)\oplus (C\otimes V_{\top})=
(I\otimes -C)\oplus C=R.$$
Consequently, $C\leq_k C\otimes \neg I$ (by definition of $\mbox{diff}(R,I)$).
Hence, $C\otimes I\leq_k C\otimes \neg I\otimes I=V_{\bot}$.
That is, $C\otimes I=V_{\bot}$.
Since $C'\leq C$, it follows that $C'\otimes I=V_{\bot}$.
We have:
$I\otimes -C\leq_k R=(B\otimes -C')\oplus C'$.
Thus,
$$I\otimes -C=
(I\otimes -C)\otimes I\leq_k ((B\otimes -C')\oplus C')\otimes I=
((B\otimes -C')\otimes I)\oplus (C'\otimes I)=$$
$$=((B\otimes -C')\otimes I)\oplus V_{\bot}= (B\otimes -C')\otimes I
\leq_k B\otimes -C'.$$
That is,
\begin{equation}\label{equ-3}
I\otimes -C\leq_k B\otimes -C'.
\end{equation}
Since $R$ is consistent, $C'$ is consistent, too.
It means that $C'\leq_k -C'$. Hence, $B\otimes -C'\geq_k B\otimes C'$.
Therefore,
$$R\otimes B= ((B\otimes -C')\oplus C')\otimes B =
((B\otimes -C')\otimes B)\oplus (C'\otimes B)=$$
$$= (B\otimes -C')\oplus (B\otimes C')= B\otimes -C'.$$
That is,
\begin{equation}\label{equ-4}
R\otimes B= B\otimes -C'.
\end{equation}
Similarly,
\begin{equation}\label{equ-5}
R\otimes I = I\otimes -C.
\end{equation}
Combining (\ref{equ-3}), (\ref{equ-4}), and (\ref{equ-5}) we get
$R\otimes I\leq_k R\otimes B$. \hfill$\Box$

If ${\cal T}$ is {\em not} a Boolean algebra, then the statement of the
above lemma does not necessarily hold, as illustrated by the following example.
\begin{example}{\rm 
Let ${\cal T} = {\cal T}_{[0,1]}$, $U=\{a\}$.
Let $R(a)=\lacute 0.3, 0.7\racute$, $B(a)=\lacute 0.2, 0.5\racute$,
and $I(a)=\lacute 0.1, 0.6\racute$. Clearly, $R$ and $I$ are consistent.
It is easy to see that
$(\mbox{diff}(R,B))(a) = (\mbox{diff}(R,I))(a) = \lacute 0.3, 0.7\racute$.
Hence, $\mbox{diff}(R,B)\leq_k \mbox{diff}(R,I)$.
However, $R\otimes B\not\geq_k R\otimes I$.
Indeed, $(R\otimes B)(a) = \lacute 0.2, 0.5\racute$,
and $(R\otimes I)(a) = \lacute 0.1, 0.6\racute$.
}\hfill$\Box$ 
\end{example}

\begin{theorem}\label{lessdiff}
Let ${\cal T}$ be a Boolean algebra.
Let $R$ be a consistent $P$-justified revision of a consistent $I$.
Let $C=\mbox{diff}(R,I)$.
Let $B$ be such that $\mbox{diff}(R,B)=C'\leq_k C$.
Then, $R$ is a $P$-justified revision of $B$.
\end{theorem}
Proof.
Consider two reducts $P_R|I$ and $P_R|B$.
Let $r'\in P_R$. Let $(l\cln \beta)$ be an annotated revision atom
from the body of $r'$.
Let $(\theta^{-1}(I))(l) = \delta_I$, $(\theta^{-1}(B))(l)=\delta_B$,
and $(\theta^{-1}(R))(l) = \delta_R$.
By the definition of a reduct, the corresponding rule in $P_R|I$
contains in the body the annotated revision literal $(l\cln \gamma_I)$,
where $\gamma_I=\pcomp(\delta_I,\beta)$. The corresponding rule in
$P_R|B$ contains in the body the annotated revision literal
$(l\cln \gamma_B)$, where $\gamma_B=\pcomp(\delta_B,\beta)$.
By the definition of pseudocomplement,
\begin{equation}\label{eqa-3}
\delta_I\vee \gamma_I\geq \beta.
\end{equation}
Since $r'\in P_R$, $\beta\leq \delta_R$.
Hence, $\beta\wedge \delta_R = \beta$. Also,
from the definition of $pcomp$ we get
$\gamma_I\leq \beta$, which implies $\gamma_I\wedge \delta_R = \gamma_I$.
From (\ref{eqa-3}) we get
$$(\delta_I\vee \gamma_I)\wedge \delta_R\geq \beta\wedge \delta_R.$$
That is,
$$(\delta_I\wedge \delta_R)\vee \gamma_I\geq \beta.$$
From Lemma \ref{inert} it follows that
$\delta_B\wedge \delta_R\geq \delta_I\wedge \delta_R$.
Therefore,
$$\delta_B\vee \gamma_I\geq (\delta_B\wedge \delta_R)\vee \gamma_I\geq \beta.$$
From definition of $\pcomp(\delta_B,\beta)$ it follows that
$\gamma_B\leq \gamma_I$.
This means that the only difference between reducts $P_R|I$ and $P_R|B$
is that annotations of literals in the bodies of rules from $P_R|B$ are less
than annotations of corresponding literals in $P_R|I$.
Consequently, $NC(P_R|B)\geq_k NC(P_R|I)$.

Since $R$ is consistent,
$$C'\leq_k C\leq_k NC(P_R|I)\leq_k NC(P_R|B)\leq_k R\leq_k$$
$$\leq_k-R\leq_k -NC(P_R|B)
\leq_k -C\leq_k -C'.$$
Also, $R=(B\otimes -C')\oplus C'$ implies
that $B\otimes -C'\leq_k R$, and $B\oplus C'\geq_k R$.
Then, on one hand,
$$(B\otimes -NC(P_R|B))\oplus NC(P_R|B)\leq_k (B\otimes -C')\oplus R
\leq_k R\oplus R=R.$$
On the other hand,
$$(B\otimes -NC(P_R|B))\oplus NC(P_R|B)=
(B\oplus NC(P_R|B))\otimes -NC(P_R|B)\geq_k$$
$$\geq_k(B\oplus C')\otimes R\geq_k
R\otimes R=R.$$
Therefore, $(B\otimes -NC(P_R|B))\oplus NC(P_R|B)=R$.
That is, $R$ is a $P$-justified revision of $B$.
\hfill$\Box$

\begin{theorem}
Let ${\cal T}$ be a Boolean algebra.
Let \ $R$ \ be a consistent \ $P$-justified revision of a consistent $I$.
Then, \ $\mbox{diff}(R,I)$ \ is minimal in the family
$\{\mbox{diff}(B,I) :  B$ is a consistent model of $P\}$.
\end{theorem}
Proof.
Let $C=\mbox{diff}(R,I)$. Then,
$R=(I\otimes -C)\oplus C$.
Since $R$ is consistent, $C$ is also consistent.
That is, $C\leq_k -C$.
Let $B$ be a consistent model of $P$, and let $\mbox{diff}(B,I) = C'\leq_k C$.
We have $B=(I\otimes -C')\oplus C'$.
Inequality $C'\leq_k C$ implies $C'\leq_k C\leq_k-C\leq_k -C'$.
Therefore,
$$(B\otimes -C)\oplus C = (((I\otimes -C')\oplus C')\otimes -C)\oplus C =$$
$$=(I\otimes -C'\otimes -C)\oplus (C'\otimes -C)\oplus C =
(I\otimes -C)\oplus C'\oplus C =$$
$$=(I\otimes -C)\oplus C = R.$$
Consequently, $\mbox{diff}(R,B)\leq_k C$.
By Theorem \ref{lessdiff},
$R$ is a $P$-justified revision of $B$.
However, $B$ is a consistent model of $P$. By Theorem \ref{OnlyModel},
$B$ is the only $P$-justified revision of itself.
Therefore, $R=B$.    \hfill$\Box$

The condition in the above theorem that revision is
consistent is important.
For inconsistent revisions the minimality principle does not hold,
as shown in the following example.

\begin{example}{\rm 
Let ${\cal T} = {\cal T}_{\{p\}}$ with the De Morgan complement being
the set-theoretic complement.
Let $P$ be an annotated revision program consisting of the following rules:
\begin{eqnarray*}
(\inn(a)\cln \{p\}) & \leftarrow & \\
(\out(a)\cln \{p\}) & \leftarrow & (\out(a)\cln \{p\})
\end{eqnarray*}
Let $I(a) = \lacute\emptyset,\{p\}\racute$. Clearly, $I$ is consistent.
Let $R_1(a) =\lacute\{p\},\{p\}\racute$ and
$R_2(a) =\lacute\{p\},\emptyset\racute$.
Both $R_1$ and $R_2$ are $P$-justified revisions of $I$.
Thus, $R_1$ is inconsistent s-model of $P$, and $R_2$ is consistent
model of $P$.
We have: $\mbox{diff}(R_1,I)= \lacute\{p\},\{p\}\racute$, and
$\mbox{diff}(R_2,I)= \lacute\{p\},\emptyset\racute$.
Clearly, $\mbox{diff}(R_2,I)\leq_k \mbox{diff}(R_1,I)$.
Therefore, $R_1$ is a $P$-justified revision of a consistent $I$,
but $\mbox{diff}(R_1,I)$ is {\em not} minimal in the family
$\{\mbox{diff}(B,I) :  B$ is a consistent model of $P\}$.
}\hfill$\Box$ 
\end{example}

\section{An alternative way of describing annotated revision programs and
order isomorphism theorem}\label{shift}
 
We will now provide an alternative description of annotated revision
programs. Instead of evaluating separately {\em revision} atoms in $\cal T$ 
we will evaluate atoms in ${\cal T}^2$. This
alternative presentation will allow us to obtain a result on the
preservation of justified revisions under order isomorphisms 
of ${\cal T}^2$.
This result is a generalization of
the ``shifting theorem'' of \cite{mpt98}.
 
An expression of the form $a\cln\langle\alpha ,\beta \rangle$, where
$\langle\alpha ,\beta \rangle\in {{\cal T}^2}$, will be called an {\em
annotated atom} (thus, annotated atoms are {\em not} annotated revision
atoms). Intuitively, an atom $a\cln\langle\alpha ,\beta \rangle$ stands
for  the conjunction of $(\inn (a)\cln \alpha)$ and $(\out (a)\cln \beta)$.
An {\em annotated rule} is an expression of the form
$p\leftarrow q_1,\dots,q_n$ where $p, q_1,\dots,q_n$ are
annotated atoms. An {\em annotated program} is a set of annotated rules.
 
A ${{\cal T}^2}$-valuation $B$ {\em satisfies} an annotated 
atom $a\cln\langle\alpha,\beta \rangle$ 
if $\langle\alpha ,\beta \rangle \leq_k B(a)$. This
notion of satisfaction can be extended to annotated rules and
annotated programs.
 
We will now define the notions of reduct, necessary change and
justified revision for the new kind of programs. 
Let $P$ be an annotated program. 
Let $B_I$ and $B_R$ be two ${{\cal T}^2}$-valuations.
The reduct of a program
$P$ with respect to two valuations $B_I$ and $B_R$ is defined in a
manner similar to Definition \ref{defreduct}. Specifically, we leave only
the rules with bodies that are satisfied by $B_R$, and in the remaining rules we
reduce the annotated atoms (except that now the transformation $\theta$
is no longer needed!). 
 
\begin{definition}\label{nsdefreduct}
The {\em reduct} ${P}_{B_R}|B_I$ is obtained from $P$ by
\begin{enumerate}
\item removing every rule whose body contains an annotated
atom that is not satisfied in $B_R$,
\item replacing each annotated atom $l\cln \beta$
from the body of each remaining rule by the annotated atom
$l\cln \gamma$, where $\gamma = \pcomp (B_I(l),\beta)$ 
(here $\beta, \gamma \in {{\cal T}^2}$).
\end{enumerate}
\end{definition}

Next, we compute the least fixpoint of the
operator associated with the reduced program. Finally, as in
Definition \ref{defjr},
we define the concept of justified revision of a valuation $B_I$ with
respect to a revision program $P$.
 
\begin{definition}\label{nsdefjr}
$B_R$ is a ${P}$-{\em justified revision} of $B_I$ if
$B_R = (B_I\otimes -C)\oplus C$, where $C=NC(P_{B_R}|B_I)$ is the
necessary change for ${P}_{B_R}|B_I$.
\end{definition}

It turns out that this new syntax does not lead to a new notion of
justified revision. Since we talk about two different syntaxes, we will
use the term ``old syntax'' to denote the revision programs as defined
in Section \ref{prelim}, and ``new syntax'' to describe programs
introduced in this section.
Specifically we now exhibit two mappings. The first of them, $tr_1$,
assigns to each ``old'' in-rule
\[
(\inn(a)\cln \alpha) \leftarrow
(\inn(b_1)\cln \alpha_1),\dots,(\inn(b_m)\cln \alpha_m),
(\out(s_1)\cln \beta_1),\dots,(\out(s_n)\cln \beta_n),
\]
a ``new'' rule
\[
a\cln\lacute\alpha,\bot\racute \leftarrow
b_1\cln\lacute\alpha_1,\bot\racute,\dots,b_m\cln\lacute\alpha_m,\bot\racute,
s_1\cln\lacute\bot,\beta_1\racute,\dots,s_n\cln\lacute\bot,\beta_n\racute.
\]
An ``old'' out-rule
\[
(\out(a)\cln \beta) \leftarrow
(\inn(b_1)\cln \alpha_1),\dots,(\inn(b_m)\cln \alpha_m),
(\out(s_1)\cln \beta_1),\dots,(\out(s_n)\cln \beta_n)
\]
is encoded in analogous way:
\[
a\cln\lacute\bot,\beta\racute \leftarrow
b_1\cln\lacute\alpha_1,\bot\racute,\dots,b_m\cln\lacute\alpha_m,\bot\racute,
s_1\cln\lacute\bot,\beta_1\racute,\dots,s_n\cln\lacute\bot,\beta_n\racute.
\]
Translation $tr_2$, in the other direction, replaces a ``new" revision rule 
by one in-rule and one out-rule. Specifically, a ``new'' rule
\[
a\cln\langle \alpha ,\beta \rangle \lar a_1\cln\langle \alpha_1 ,\beta_1
\rangle,\ldots,
a_n\cln\langle \alpha_n ,\beta_n \rangle
\]
is replaced by two ``old'' rules (with identical
bodies but different heads)
\[
(\inn (a)\cln \alpha) \lar (\inn (a_1)\cln \alpha_1), (\out (a) \cln \beta_1),
\ldots ,
(\inn (a_n)\cln \alpha_n) , (\out (a_n) \cln \beta_n)
\]
and
\[
(\out (a)\cln \beta) \lar (\inn (a_1)\cln \alpha_1), (\out (a) \cln \beta_1),\ldots
 ,
(\inn (a_n)\cln \alpha_n) , (\out (a_n) \cln \beta_n).
\]
The translations $tr_1$ and $tr_2$ can be extended to programs.
We then have the following theorem that states that the new syntax and
semantics of annotated revision programs presented in this section are 
equivalent to the syntax and semantics introduced and studied earlier 
in the paper.
 
\begin{theorem}
Both transformations $tr_1$, and $tr_2$ preserve justified revisions.
That is, if $B_I, B_R$ are valuations in ${\cal T}^2$
and $P$ is a program in the ``old''
syntax, then $B_R$ is a $P$-justified revision of $B_I$ if and only if
$B_R$ is a $tr_1 (P)$-justified revision of $B_I$. Similarly,
if $B_I, B_R$ are  valuations in ${\cal T}^2$  and $P$ is a program
in the ``new''
syntax, then $B_R$ is a $P$-justified revision of $B_I$ if and only if
$B_R$ is a $tr_2 (P)$-justified revision of $B_I$.
\end{theorem}
 
In the case of unannotated revision programs, the shifting theorem
proved in \cite{mpt98} shows that for every revision program $P$ and
every two initial databases $B$ and $B'$ there is a revision program
$P'$ such that there is a one-to-one correspondence between
$P$-justified revisions of $B$ and $P'$-justified revisions of $B'$.
In particular, it follows that the study of justified revisions 
(for unannotated programs) can
be reduced to the study of justified revisions of empty databases.
We will now present a counterpart of this result for annotated revision
programs. The situation here is more complex. It is no longer true that
a ${\cal T}^2$-valuation can be ``shifted'' to any other ${\cal
T}^2$-valuation. However, the shift is possible if the two valuations are
related to each other by an order isomorphism of the lattice of all
${\cal T}^2$-valuations. 
 
There are many examples of order isomorphisms on the lattice of
${\cal T}^2$. For instance,
the mapping $\psi:{\cal T}^2\rightarrow{\cal T}^2$ defined by
$\psi(\langle\alpha,\beta\rangle) = \langle \beta,\alpha\rangle$
is an order isomorphism of ${\cal T}^2$. In the case of the
lattice ${\cal T}_X$, order isomorphisms of ${\cal T}^2_X$
can also be generated by permutations of the set $X$. 

Let $\psi$ be an order isomorphism on ${\cal T}^2$.
It can be extended to annotated atoms, annotated rules, and
${\cal T}^2$-valuations as follows:  \\
$\psi(a:\delta) = a:\psi(\delta)$, \\
$\psi(a\cln\delta \lar a_1\cln\delta_1,\ldots, a_n\cln\delta_n) = 
\psi(a\cln\delta)\lar \psi(a_1\cln\delta_1),\ldots, \psi(a_n\cln\delta_n)$, \\
$(\psi(B))(a) = \psi(B(a))$, \\
where $a,a_1,\dots,a_n\in U$, $\delta,\delta_1,\dots,\delta_n\in {\cal T}^2$,
and $B$ is a ${\cal T}^2$-valuation.

The extension of an order isomorphism on ${\cal T}^2$ 
to ${\cal T}^2$-valuations is again an order isomorphism,
this time on the lattice of all ${\cal T}^2$-valuations.
We say that an order isomorphism $\psi$ on a lattice {\em preserves 
conflation}
if $\psi(-\delta)=-\psi(\delta)$ for all elements $\delta$ from the lattice.
We now have the following result that generalizes the shifting theorem of
\cite{mpt98}.
 
\begin{theorem}\label{OrderIsom}
Let $\psi$ be an order isomorphism on the set of ${\cal T}^2$-valuations.
Let $\psi$ preserve conflation.
Then, $B_R$ is a ${P}$-justified revision of $B_I$ if and only if
$\psi(B_R)$ is a $\psi({P})$-justified revision of $\psi(B_I)$.
\end{theorem}
Proof.
By definition, $B_R$ is a ${P}$-justified revision of $B_I$ if and only if
$B_R = (B_I\otimes -C)\oplus C$, where $C = NC(P_{B_R}|B_I)$.
Since $\psi$ is an order isomorphism, it preserves meet and join operations.
Therefore,
$$\psi(B_R) = \psi((B_I\otimes -C)\oplus C) = 
\psi(B_I\otimes -C)\oplus \psi(C) =$$ 
$$=(\psi(B_I)\otimes \psi(-C))\oplus \psi(C) =
(\psi(B_I)\otimes -\psi(C))\oplus \psi(C).$$
At the same time, $\psi(P_{B_R}|B_I) = (\psi(P))_{\psi(B_R)}|\psi(B_I)$,
and $NC(\psi(P_{B_R}|B_I)) = \psi(NC(P_{B_R}|B_I))$. 
Thus, $B_R$ is a ${P}$-justified revision of $B_I$ if and only if
$\psi(B_R)$ is a $\psi({P})$-justified revision of $\psi(B_I)$.
\hfill $\Box$

Shifting theorem of \cite{mpt98}, that applies to ordinary revision
programs, is just a particular case of Theorem \ref{OrderIsom}. 
In order to derive it from Theorem \ref{OrderIsom}, we take 
${\cal T}={\cal TWO}$. Next, we consider an ordinary revision program 
$P$ and two databases $B_1$ and $B_2$ (let us recall that in the case 
of ordinary revision programs, databases are {\em sets of atoms} and 
not valuations). Let $P^a$ and $B_1^v$ and $B_2^v$ be defined as in 
Theorem \ref{a-2}. It is easy to see that the operator $\psi$, defined
by
$$
(\psi(v))(a) =
\cases{\langle \beta,\alpha\rangle, & when $B_1^v(a)\not = B_2^v(a)$\cr
\langle\alpha,\beta\rangle, & when $B_1^v(a)=B_2^v(a)$},
$$
is an order-isomorphism on ${\cal TWO}^2$-valuations and that
$\psi(B_1^v) = B_2^v$. Let $C_1$ and $C_2$ be two sets of atoms such 
that $C_2^v = \psi(C_1^v)$. By Theorem \ref{OrderIsom}, $C_1^v$ is a 
$P^a$-justified revision of $B_1^v$ if and only if $C_2^v$ is a
$\psi(P^a)$-justified revision of $B_2^v$. Theorem \ref{a-2} and the
observation that the necessary change of $P^a_{C_1^v}|B_1^v$ is
consistent if and only if the necessary change of
$\psi(P^a)_{C_2^v}|B_2^v$ is consistent together imply now the shifting 
theorem of \cite{mpt98}.

The requirement in Theorem \ref{OrderIsom} that $\psi$ preserves 
conflation is essential. If it is not the case, the statement of the 
theorem may not hold as illustrated by the following example.

\begin{example}{\rm 
Let ${\cal T} = {\cal T}_{\{p,q,r\}}$ with the De Morgan complement defined
as follows:
\begin{eqnarray*}
\overline{\{\}}=\{p,q,r\}, \quad & \overline{\{p\}}=\{p,r\}, \quad 
\overline{\{q\}}=\{q,r\}, \ \quad & \overline{\{r\}}=\{p,q\}, \\
\overline{\{p,q,r\}}=\{\}, \ \quad & \overline{\{p,r\}}=\{p\}, \quad 
\overline{\{q,r\}}=\{q\}, \ \quad & \overline{\{p,q\}}=\{r\}.
\end{eqnarray*}
Let $\psi$ be order isomorphism on ${\cal T}$ such that
$\psi(\{p\})=\{p\}$, $\psi(\{q\})=\{r\}$, and $\psi(\{r\})=\{q\}$.
Clearly, $\psi$ does not preserve conflation, because
$$\psi(-\lacute\{p\},\{\}\racute) = \psi(\lacute\{p,q,r\},\{p,r\}\racute) = 
\lacute\{p,q,r\},\{p,q\}\racute, \mbox{ but}$$
$$-\psi(\lacute\{p\},\{\}\racute) = -\lacute\{p\},\{\}\racute = 
\lacute\{p,q,r\},\{p,r\}\racute.$$
Let an annotated program be the following:
\begin{eqnarray*}
P: \quad \quad a:\lacute\{p\}, \{\}\racute & \leftarrow
\end{eqnarray*}
It determines the necessary change $C(a)=\lacute\{p\},\{\}\racute$.
 
Then, $-C(a)=\lacute\{p,q,r\},\{p,r\}\racute$.
Let $B_I(a) = \lacute\{\},\{r\}\racute$.
The $P$-justified revision of $B_I$ is
$B_R(a)=(\lacute\{\},\{r\}\racute\otimes\lacute\{p,q,r\},\{p,r\}\racute)\oplus
\lacute\{p\},\{\}\racute = \lacute\{p\},\{r\}\racute$.
 
The annotated program $\psi(P)$ is the same as $P$.
We have $\psi(B_I)(a)=\lacute\{\},\{q\}\racute$, 
$\psi(B_R)(a)=\lacute\{p\},\{q\}\racute$.
The reduct $(\psi(P))_{\psi(B_R)}|\psi(B_I) = \psi(P) = P$.
The necessary change determined by the reduct is $C$.
However,
$$((\psi(B_I)\otimes -C)\oplus C)(a) = \lacute\{p\},\{\}\racute \not=
\psi(B_R)(a).$$
Therefore, $\psi(B_R)$ is {\em not} a $\psi(P)$-justified revision
of $\psi(B_I)$.
}\hfill$\Box$
\end{example}

\section{Conclusions and further research}\label{concl}
 
The main contribution of our paper is a new definition of the reduct
(and hence of a justified revision) for annotated programs considered
by Fitting in \cite{fit95}. This new definition eliminates some 
anomalies arising in the approach by Fitting. Specifically, in Fitting's
approach, justified revisions are not, in general, models of a program. In
addition, they do not satisfy the invariance-under-join property. In our
approach, both properties hold. Moreover, as we show in Sections
\ref{props} and \ref{shift}, many key properties of ordinary revision 
programs
extend to the case of annotated revision programs under our definition
of justified revisions. 
 
Several research topics need to be further pursued. First, the concepts 
of an annotated revision program and of a justified revision can be 
generalized to the disjunctive case, where a program may have 
``nonstandard disjunctions'' in the head. One can show that this
extension indeed reduces back to the ordinary concept of annotated
revision programming, as discussed here, if no rule of a program
contains a disjunction in its head. However, an in-depth study of
annotated disjunctive revision programming has yet to be conducted.

Second, in this paper we focused on the case when the lattice of 
annotations is distributive. This assumption can be dropped and a
reasonable notion of a justified revision can still be defined.
However, the corresponding theory is so far less understood and it seems
to be much less regular than the one studied in this paper.

Finally, we did not study here the complexity of reasoning tasks for
annotated revision programs. Assuming that the lattice is finite and
fixed (is not part of the input), the complexity results obtained in
\cite{mt94c} can be extended to the annotated case. The complexity of
reasoning tasks when the lattice of annotations is a part of an input
still needs to be studied. Clearly, any such study would have to take
into account the complexity of evaluating lattice operations.

\section{Acknowledgments}
This work was partially supported by the NSF grants CDA-9502645 and 
IRI-9619233.

\newcommand{\etalchar}[1]{$^{#1}$}

\end{document}